\documentclass[lettersize,journal]{IEEEtran}
\usepackage{amsmath,amsfonts}
\usepackage{algorithmic}
\usepackage{algorithm}
\usepackage{array}
\usepackage{textcomp}
\usepackage{stfloats}
\usepackage{url}
\usepackage{verbatim}
\usepackage{graphicx}
\usepackage{cite}
\usepackage{subfigure}
\usepackage{mathtools}
\usepackage{float}
\usepackage[export]{adjustbox}
\usepackage{arydshln}
\usepackage{tensor}
\usepackage{multirow}
\usepackage{booktabs} 
\usepackage{gensymb}
\usepackage{bbding}
\usepackage{pifont}
\usepackage{bm}
\usepackage{subcaption}
\usepackage{siunitx}
\usepackage[colorlinks,linkcolor=black]{hyperref}
\usepackage[switch]{lineno}
\usepackage{relsize}
\usepackage{threeparttable}

\newcommand{\figref}[1]{Fig.~\ref{#1}}
\newcommand{\tabref}[1]{Table.~\ref{#1}}
\newcommand{\equref}[1]{Eqn.~\ref{#1}}
\newcommand{\secref}[1]{Sec.~\ref{#1}}

\newenvironment{nocolor}{%
    \renewcommand{\textcolor}[2]{##2}%
}{}  

\begin{document}
\begin{nocolor}

\title{\textcolor{red}{An Online Adaptation Method for Robust Depth Estimation and Visual Odometry in the Open World}}
\author{Xingwu Ji, 
        Haochen Niu, Dexin Duan,
        Rendong Ying, 
        Fei Wen, 
        Peilin Liu 
\thanks{The authors are with the Brain-inspired Application Technology Center (BATC), School of Electronic Information and Electrical Engineering, Shanghai Jiao Tong University, Shanghai 200240, China (email: \{jixingwu, haochen\_niu, jumpywizard, rdying, wenfei, liupeilin\}@sjtu.edu.cn).}}



\maketitle

\begin{abstract}
Recently, learning-based robotic navigation systems have gained extensive research attention and made significant progress. However, the diversity of \textcolor{red}{open-world} scenarios poses a major challenge for the generalization of such systems to practical scenarios. \textcolor{red}{Specifically, learned systems for scene measurement and state estimation tend to degrade when the application scenarios deviate from the training data, resulting to unreliable depth and pose estimation.} Toward addressing this problem, this work aims to develop a \textcolor{red}{visual odometry} system that can fast adapt to diverse novel environments in an online manner. To this end, we construct a self-supervised online adaptation framework for monocular \textcolor{red}{visual odometry} aided by an online-updated depth estimation module. Firstly, we design a monocular depth estimation network with lightweight refiner modules, which enables efficient online adaptation. Then, we construct an objective for self-supervised learning of the depth estimation module based on the output of the \textcolor{red}{visual odometry} system and the contextual semantic information of the scene. Specifically, a sparse depth densification module and a dynamic consistency enhancement module are proposed to leverage camera poses and contextual semantics to generate pseudo-depths and valid masks for the online \textcolor{red}{adaptation}. Finally, we demonstrate the robustness and generalization capability of the proposed method in comparison with state-of-the-art learning-based approaches on urban, in-house datasets and a robot platform. Code is publicly available at: \url{https://github.com/jixingwu/SOL-SLAM}.
\end{abstract}


\section{Introduction}
\IEEEPARstart{V}{isual} \textcolor{red}{odometry (VO) or simultaneous localization and mapping (SLAM)} plays an important role in the field of robotic navigation for environment perception \cite{liu2023unsupervised} and motion estimation \cite{wang2024unsupervised}. Geometry-based SLAM methods \cite{engel2017direct, campos2021ORBSLAM3} have shown remarkable performance under restrictive conditions, such as abundant features and static assumptions. Recently, learning-based \textcolor{red}{SLAM/VO} methods \cite{fu2024islam, xiong2021self, li2021deepslam} have gained increasing attention due to their capability to leverage learnable features. Typically, such methods design image reconstruction losses to jointly learn camera poses in a self-supervised manner, yielding promising results. However, these methods are trained on data from a specific domain, \textcolor{red}{and tested within the same setting and visual measurement}. For instance, the works \cite{zhao2020towards, wang2019improving, yin2018geonet, li2018undeepvo, wang2024unsupervised} are pre-trained on sequences 00 to 08 of the KITTI dataset \cite{geiger2012kitti} using supervised or self-supervised methods, and then tested on sequence 09 and 10. \textcolor{red}{In contrast to controlled environments where training and testing data have similar measurement distributions, real-world applications require systems to operate in open-world scenarios with domain shifts, such as unknown camera intrinsics and varying visual measurements. When these methods are deployed in such conditions, they often exhibit subtantial performance degradation and uncertainty increases in depth and pose estimation.}

\begin{figure}[tp]
    \centering
    \subfigure[\textcolor{red}{Depth estimation results}]{
    \includegraphics[width=0.46\linewidth]{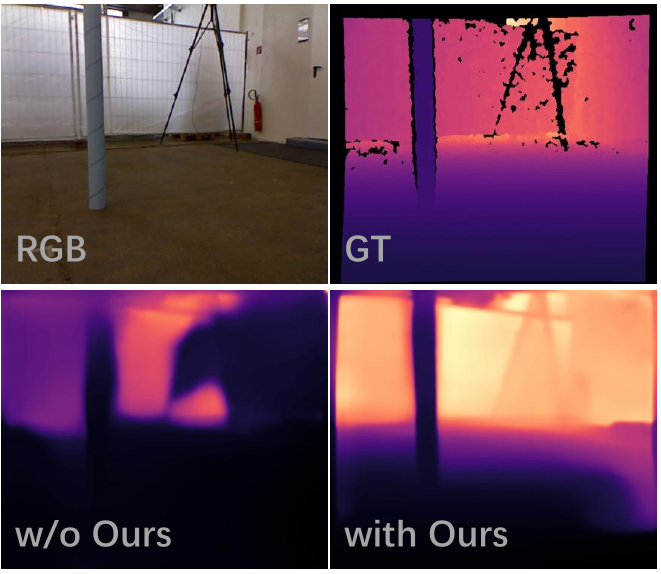}}
    \subfigure[\textcolor{red}{Visual odometry results}]{
    \includegraphics[width=0.47\linewidth]{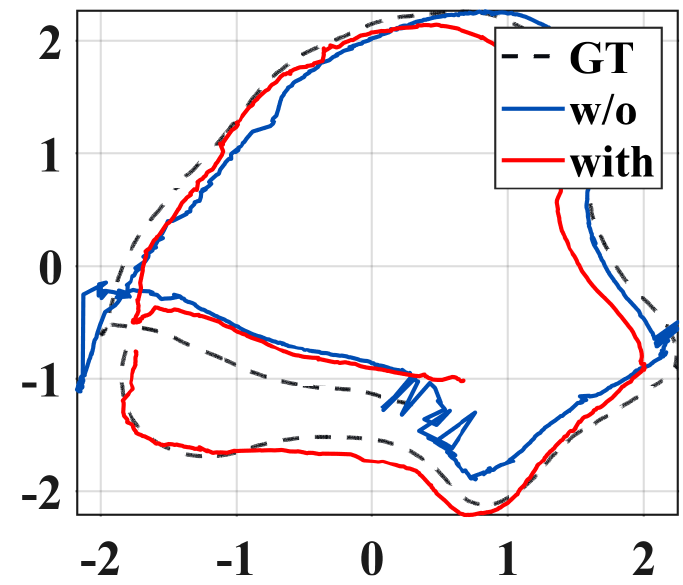}}
    \caption{An example of our proposed method on TUM dataset.}
    \label{fig:firstfigure}
\end{figure}

To improve generalization \textcolor{red}{in the open world}, adaptive \textcolor{red}{VO} methods have been developed \cite{zhang2020online, li2021generalizing, vodisch2023covio, vodisch2022continual, xu2023online}, which aim to generalize pre-trained models to new environments to enhance the robustness during testing. The works \cite{zhang2020online, li2021generalizing} utilize fine-tuning strategies to optimize the entire pre-trained models. Though effective, these works come with substantial computational requirements that may affect real-time performance, which is critical for practical applications. They also confront the challenge of catastrophic forgetting, where the models tend to lose previously acquired knowledge from source domains when adapting to new environments. To avoid the catastrophic forgetting problem, the works \cite{vodisch2023covio, vodisch2022continual} incorporate experience replay mechanisms to periodically reintroduce past experiences from source domains into the \textcolor{red}{adaptation} process. When the reintroduction of past experiences does not align seamlessly with the current learning objectives, they would suffer from slow convergence. Moreover, the work \cite{xu2023online} integrates adapters into the network architecture and only optimizes the adapters during testing. 
However, these methods do not consider \textcolor{red}{adaptation process} and SLAM systems in a holistic manner, and hence cannot make use of the pose estimation and \textcolor{red}{scene measurement} capabilities of the SLAM system to \textcolor{red}{facilitate robust and fast adaptation}.

To address these problems, this work introduces an online \textcolor{red}{adaptation} framework for {monocular} \textcolor{red}{depth estimation and VO} based on a {self-supervised} mechanism. When applied to real-world scenarios, the framework can fast adapt to diverse novel environments. 
Firstly, we design a monocular depth estimation network with lightweight refiners, namely R-DepthNet. The refiner is a low-rank matrix module with a small set of parameters. It can refine feature representations of each layer and transfer them to new environments without prior training. In the online \textcolor{red}{adaptation} stage, we freeze the pre-trained weights of DepthNet and only optimize the parameters of the refiners. 

Then, we propose a closed-loop {self-supervised} mechanism, which allows R-DepthNet and {monocular} SLAM to reinforce each other. 
Specifically, we propose a Sparse Depth Densification (SDD) module to generate pseudo-depth \textcolor{red}{measurements} based on the sparse maps of the SLAM, and a Dynamic Consistency Enhancement (DCE) module to compute valid masks based on the relative poses of the SLAM. These pseudo-depths and valid masks are used as feedback to enhance the online \textcolor{red}{adaptation process}. The learned R-DepthNet can further refine the accuracy of the SLAM output.
Finally, we conduct extensive experiments on the KITTI, TUM datasets and a mobile robot platform to demonstrate the robustness and generalization capability of our system. \figref{fig:firstfigure} shows an example of depth and pose estimation results with and without our proposed method.

Our main contributions are summarized as follows:

\begin{itemize}
     \item \textcolor{red}{An online adaptation framework for VO and depth estimation}, which can fast adapt to novel environments in an online manner based on a self-supervised mechanism.

    \item \textcolor{red}{A self-supervised online adaptation method}, in which an objective for \textcolor{red}{online adaptation process} is designed based on the output of the SLAM system and the contextual semantic information of the scene. 

    \item A monocular depth estimation network with lightweight refiners is designed for efficient online \textcolor{red}{adaptation}, namely R-DepthNet, where the refiner can transfer feature representations to new environments without prior training. 

    \item Extensive experiments on the KITTI, TUM datasets and a mobile robot platform demonstrate the effectiveness and efficiency of our proposed system. The results highlight its capability of fast generalization to novel real-world scenarios.
    
    
\end{itemize}

\section{Related Works}
\subsection{Learning-Based Visual SLAM \textcolor{red}{and VO}}
Conventional visual SLAM methods use either feature matching \cite{campos2021ORBSLAM3} or direct image alignment \cite{engel2017direct} for camera tracking and scene mapping. Recently, learning-based visual SLAM methods have been widely studied, which leverage deep learning techniques to enhance accuracy and robustness. For example, CNN-SLAM \cite{tateno2017cnn} presents a real-time dense monocular SLAM that incorporates learned depth prediction from convolutional neural networks (CNN). \textcolor{red}{In addition to monocular images, Miao et al. \cite{miao2023pseudo} propose a pseudo-LiDAR-based VO framework that leverages 3D point cloud representations derived from stereo images. Meanwhile, these works}~\cite{wang2017deepvo, xu2021attention} propose attention mechanisms within the recurrent neural networks (RNN) to extract spatial and temporal features for pose estimation.
Though promising performance has been shown, these methods require ground-truth data \textcolor{red}{for pose and depth measurements}, which may be difficult to collect in real-world scenarios.

To address this limitation, self-supervised visual SLAM \textcolor{red}{and VO} methods have been introduced \cite{li2019sequential, xiong2021self, ranjan2019competitive, wang2022motionhint}, which leverage the inherent structure within the visual \textcolor{red}{measurement} to learn a model of the environment and camera motion without external supervision signals. For example, SfMLearner \cite{zhou2017unsupervised} introduces an end-to-end learning framework to jointly estimate depth and ego-motion from monocular videos, which achieves competitive performance with supervised methods. Building on this, Monodepth2 \cite{godard2019digging} proposes enhancement including reprojection loss minimization, multi-scale sampling, and auto-masking to improve depth estimation quality. Following this idea, Zhao et al. \cite{zhao2020towards} and Ranjan et al. \cite{ranjan2019competitive} propose to jointly estimate depth, pose and optical flow. Besides, D3VO \cite{yang2020d3vo} incorporates stereo video into the training process of depths and poses to enhance the spatial perception of the system. Considering the scale ambiguity problem, Bian et al. \cite{bian2021unsupervised} propose geometric consistency constraints to achieve scale consistency of depth and pose estimation. 

However, these self-supervised methods also face the challenge of environment changes \textcolor{red}{in the open world}. The accuracy of the networks heavily depends on the visual \textcolor{red}{measurement} similarity between the training and testing data. Therefore, in the unseen environment, the performance of the methods degrades significantly.

\subsection{Domain Adaptation of \textcolor{red}{Depth Prediction and VO}}

Learning-based methods typically rely on data from specific domains or particular types of scenes. In other words, these methods \cite{zhao2020towards, wang2019improving, yin2018geonet, bian2021unsupervised} are pre-trained under the assumption that both the training and testing data are sampled from the same \textcolor{red}{visual measurement} distribution. When confronted with the unseen environments \textcolor{red}{in the open world}, they often fail to deliver optimal performance. 
To address the generalization problem in the presence of domain shift, adaptation approaches have been proposed. Such approaches enable systems to dynamically adapt to new environments, in an attempt to maintain robustness and high performance \textcolor{red}{of depth prediction \cite{gurram2022monocular, li2023test} and pose estimation \cite{pnvr2020sharingan, li2020self, pan2024adaptive, luo2018real}} in unseen scenarios. 

\textcolor{red}{For depth prediction, Gurram et al. \cite{gurram2022monocular} introduce a depth estimation network that minimizes domain discrepancies between virtual and real-world data through supervised and self-supervised learning. Building on this, Li et al. \cite{li2023test} propose 3-branch adaptation network to generate scale-aware depth predictions on targe data, leveraging pre-trained models from both supervised and self-supervised paradigms.}

For pose estimation, Pnvr et al. \cite{pnvr2020sharingan} propose to align images from source and target domains based on generative adversarial networks (GAN), ensuring stylistic and visual consistency. This alignment creates a bridge between different datasets, which can transfer learned features from one to another \cite{kundu2018adadepth}. Li et al. \cite{li2021generalizing, li2020self} introduce online adaptation frameworks that utilizes meta-learning or retraining mechanisms to fine-tune pre-trained networks \textcolor{red}{for pose and depth estimation} on target domains. Besides, Luo et al. \cite{luo2018real} accumulate images from target domains to optimize the network in the unsupervised manner. These methods continuously optimize pre-trained models to improve robustness in new environments.


Though effective, directly fine-tuning the entire network without any mechanism to preserve the knowledge from source domains suffers from the catastrophic forgetting problem. To address the problem, Xue et al. \cite{xue2020deep} design a memory module to save the information from target domains, which enables the system to process long-term dependency. Following this idea, Vodisch et al. \cite{vodisch2023covio} leverage a memory buffer to replay previous experiences during the adaptation process. These methods allow the models to revisit source domain data and maintain a balance between the source and target domains. On the other hand, Xu et al. \cite{xu2023online} integrate adapter modules into pre-trained networks, which only update the adapters for adaptation. However, this work requires pre-training complex adapters in the source domain, and is inferior to the fine-tuning and memory based methods in terms of accuracy. 

In this work, we make use of the environment agnosticism and scalability of geometry-based SLAM systems to integrate \textcolor{red}{online adaptation mechanism} into SLAM systems in a holistic manner. Specifically, we propose a closed-loop self-supervised mechanism based on R-DepthNet and the SLAM system, which can efficiently adapt to diverse novel scenes in an online manner. %

\begin{figure}[tp]
    \centering
    \includegraphics[width=0.48\textwidth]{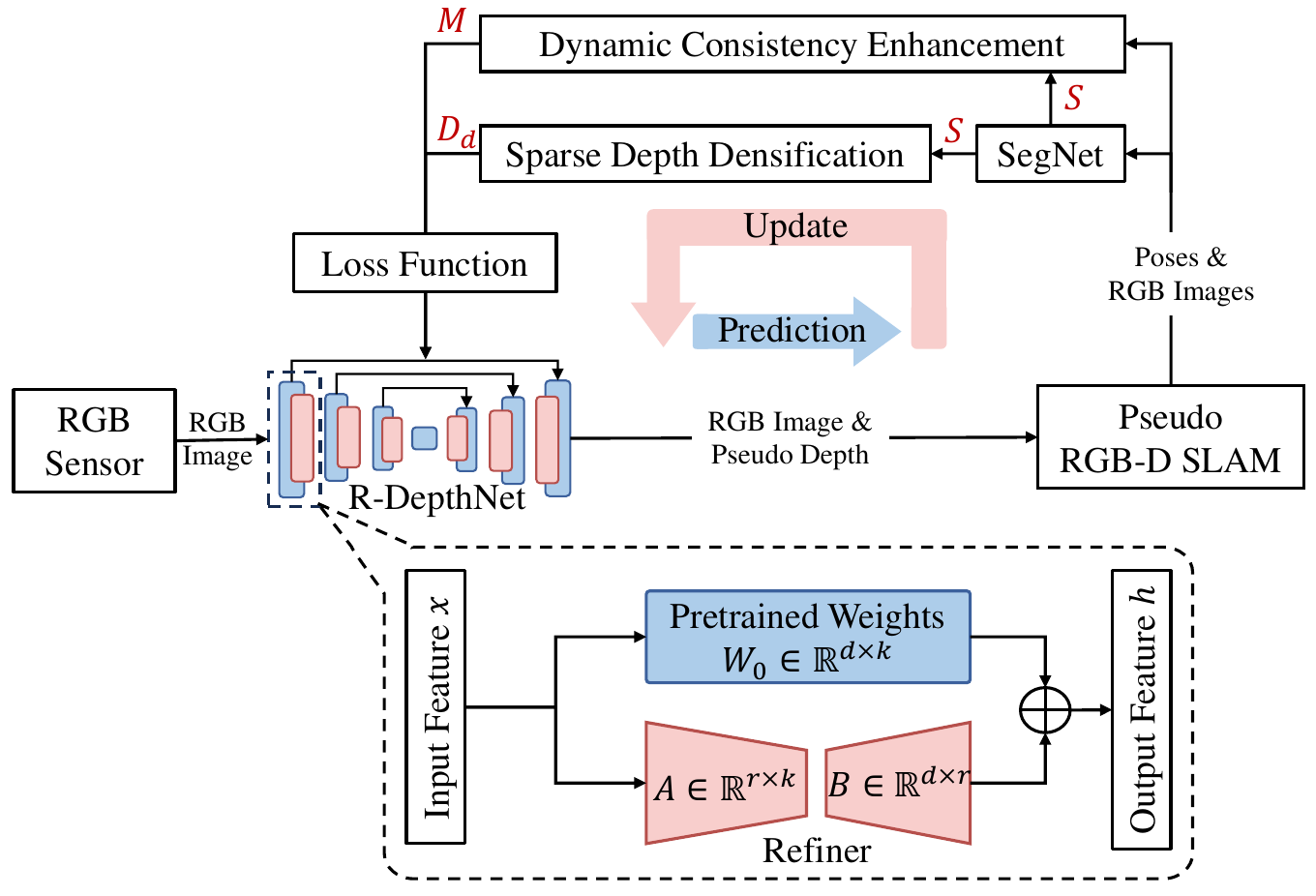}
    \caption{Overview of \textcolor{red}{our online adaptation framework}. In this system, the depth estimation module R-DepthNet and the pseudo RGB-D SLAM reinforce each other in an online manner, \textcolor{red}{and output the refined depth maps and camera poses, respectively}. $\mathcal{S}$: object regions, $M$: consistency masks and $D_d$: densified depths.}
    \label{fig:systemOverview}
\end{figure}

\section{Method}

This section introduces the framework of our system in detail. As illustrated in~\figref{fig:systemOverview}, our system  comprises two main stages, prediction and update. In the prediction stage, the R-DepthNet predicts pseudo depth maps for input images. The pseudo RGB-D SLAM then matches 2D correspondence points and calculates the relative poses. In the update stage, the SDD module computes sparse depth maps, and then densifies them into dense depth maps by gridding images and semantic objects. Moreover, the DCE module introduces a semantic and geometric consistency mask to handle moving objects. The dense depth maps and relative poses are used as feedback to construct a loss for online learning of the R-DepthNet, where the pre-trained DepthNet remains frozen while only the refiners are optimized.

\subsection{Monocular R-DepthNet}\label{sec:r-depthnet}
As illustrated in \figref{fig:systemOverview}, the R-DepthNet module is an extended version of the DepthNet \cite{bian2021unsupervised}, which incorporates lightweight refiner modules into DepthNet. DepthNet is a monocular depth estimation network, pre-trained on the source domain. The refiner modules, low-rank adapters \cite{hu2022lora}, are used for adaptation in the online learning procedure, which reduce memory requirements by utilizing a small set of trainable parameters. We freeze the pre-trained weights of DepthNet and insert the trainable adapters into each layer of the network.

Specifically, given the input feature $x$, the each layer of R-DepthNet computes the output feature $h$ as
\begin{equation}
    h = W_0x + BAx,
\end{equation}
where $W_0 \in \mathbb{R}^{d\times k}$ denotes pre-trained weights of each layer.
$B \in \mathbb{R}^{d\times r}$ and $A \in \mathbb{R}^{r\times k}$ denote the parameters of the refiner module, which are initialized as zero and random Gaussian, respectively. Here, $d$ and $k$ denote the dimensions of the input and output features, while the rank $r \ll \min(d,k)$. We only optimize $A$ and $B$ for fast adaptation in the online \textcolor{red}{manner}.

\subsection{Pseudo RGB-D SLAM}\label{sec:prgbdslam}
We employ the well-established and extensively validated visual SLAM system, ORB-SLAM3 \cite{campos2021ORBSLAM3}, to process the RGB-D data. The R-DepthNet infers the pseudo depth map for each image, which is used to match 2D correspondence points and compute camera relative poses \cite{tiwari2020pseudo, ji2022georefine}. The pseudo RGB-D SLAM system is structured with two principal phases, system initialization and bundle adjustment (BA), \textcolor{red}{while notably excluding loop closure mechanisms}. 

In the work~\cite{tiwari2020pseudo}, a virtual right correspondence is introduced, which leads to an additional reprojection error term in the BA. However, due to the inherent noise and the scale difference of depth prediction in the online learning, this approach would result in sub-optimal performance. Therefore, following~\cite{ji2022georefine}, we have omitted the reprojection error term associated with the virtual right points in our system, and take the refined depth map from the online \textcolor{red}{adaptation} as input. 

In the pipeline of the pseudo RGB-D SLAM, 3D map points are initialized using the pseudo depth maps in the first frame of the sequence. Subsequently, the system tracks the camera by matching 2D correspondence points in the local map. The relative pose is then computed with respect to minimize the reprojection error of the matched points. The system dynamically inserts keyframes and points into the local map, and enhances the accuracy of the relative poses through local BA. Notably, the pseudo depth map is utilized within the map point initialization and the new point insertion to improve the robustness of the SLAM system.

\begin{figure}[tp]
    \centering
    \subfigure[Pseudo depth (matches: 594)]{
    \includegraphics[width=0.47\linewidth]{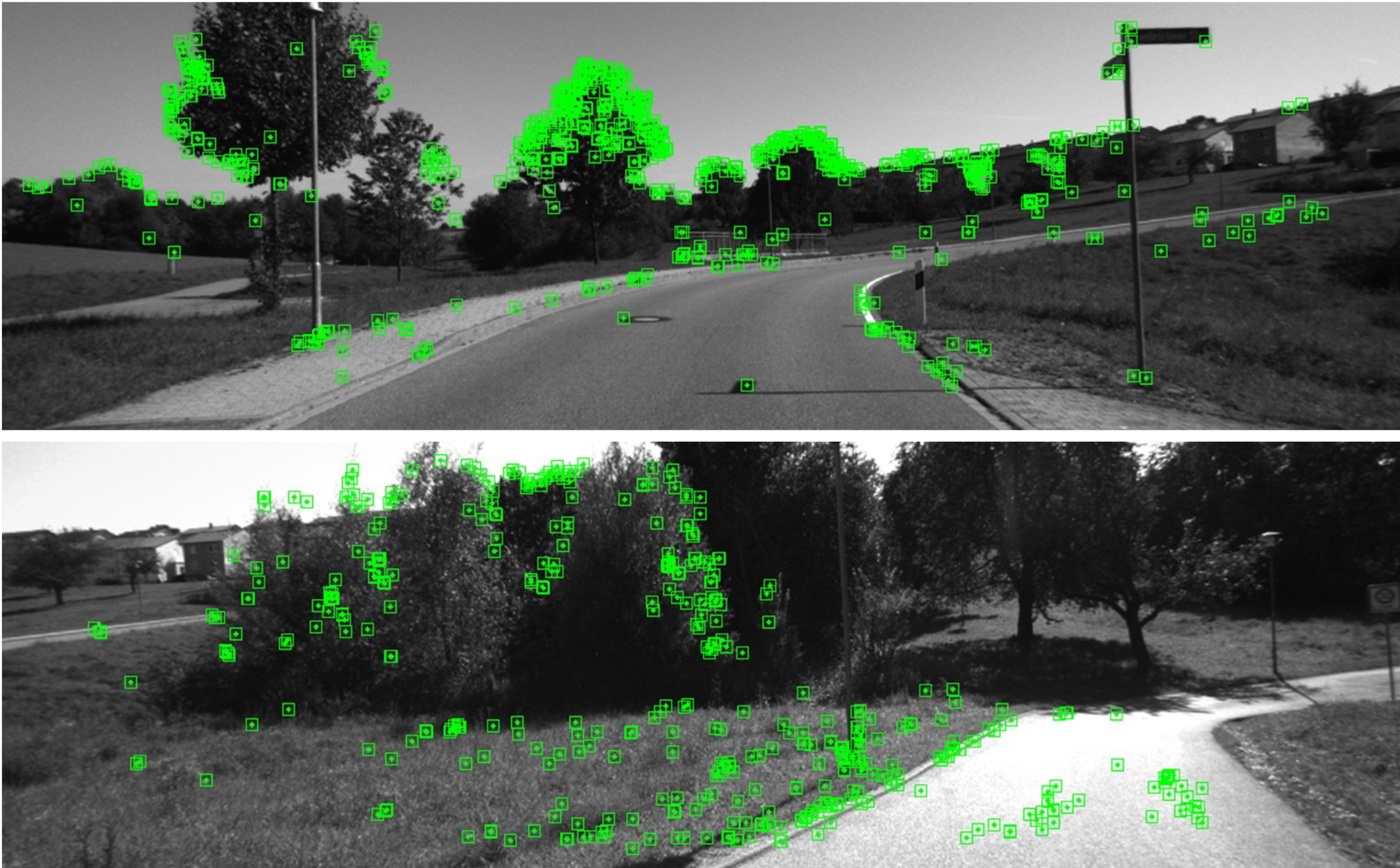}\label{fig:pseudo depth}}
    \subfigure[Triangulation (matches: 267)]{
    \includegraphics[width=0.47\linewidth]{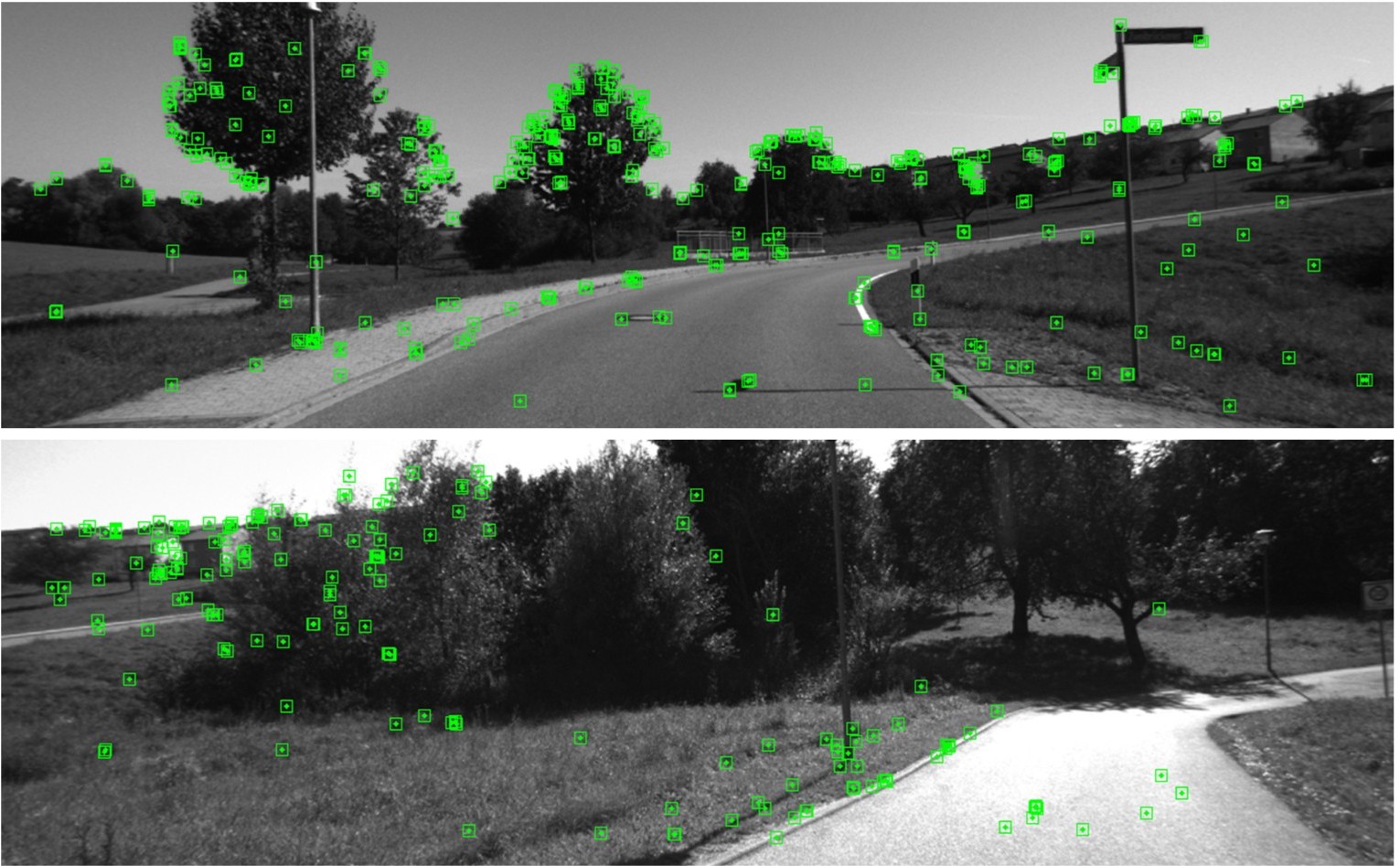}\label{fig:triangulation}}
    \caption{Extracted and matched feature points.}
\end{figure}

Specifically, the pseudo depth map is used instead of triangulation to compute feature point depth. This effectively increases the number of initialized map points, as shown in \figref{fig:pseudo depth}. In contrast, the triangulation approach often results in erroneous depth calculation due to issues such as feature point co-linearity, excessive camera motion, leading to a reduction in the number of feature points, as shown in \figref{fig:triangulation}. 

In our local BA optimization, the pseudo depth map with an associated scale $s$ is denoted as $sD_d$. The map points at the world coordinate $P_w$ and camera coordinate $P_c$ can be respectively expressed as
\begin{equation}
    \begin{aligned}
    P_w &= sD_d(p_{uv}) T^{-1}K^{-1}p_{uv} = s[X, Y, Z]^T, \\
    P_c &= TP_w = s[x, y, z]^T,
    \end{aligned}
\end{equation}
where $K$ is the camera intrinsics matrix. The residual term $e$ in the BA can be formulated as
\begin{equation}\label{equ: e}
    \begin{aligned}
        e = p_{uv} - 
        \begin{bmatrix}
        f_x \frac{x}{z} + c_x \\
        f_y \frac{y}{z} + c_y
        \end{bmatrix},
    \end{aligned}
\end{equation}
where $f_x$ and $f_y$ are the focal lengths, $(c_x, c_y)$ is the principal point.

\begin{figure}[tp]
    \centering
    \subfigure[Gridding with segmented images with feature points]{
    \label{fig:sdda}
    \includegraphics[width=0.85\linewidth]{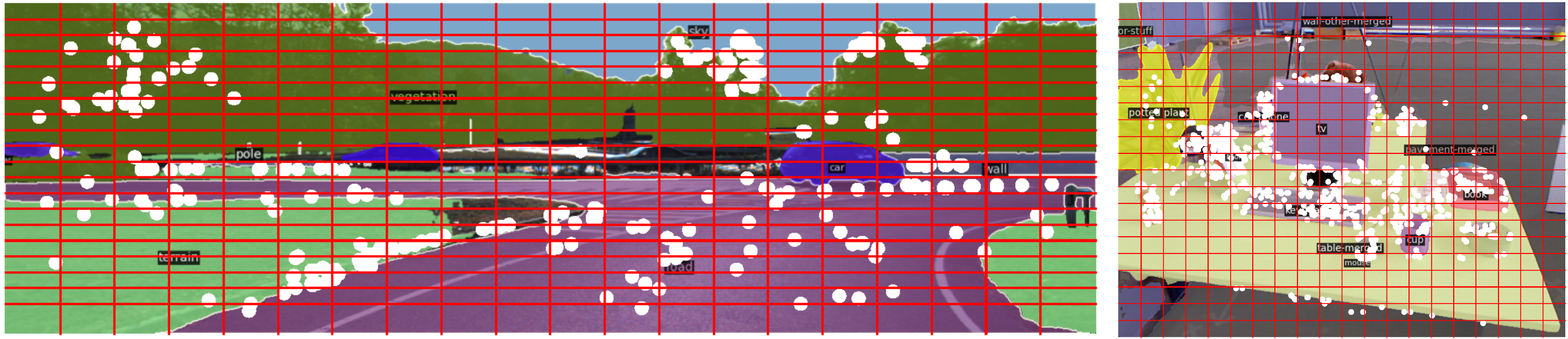}}\vspace{-1mm}
    
    \subfigure[Propagating with same categories and depth values]{
    \label{fig:sddb}
    \includegraphics[width=0.85\linewidth]{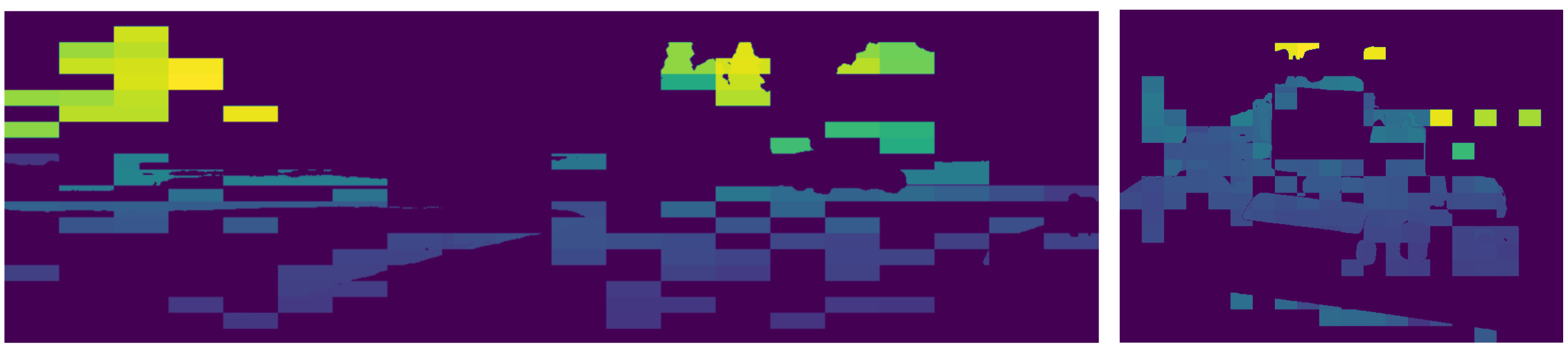}}\vspace{-1mm}
    
    \subfigure[Densifying sparse depth with semantic segmentation]{
    \label{fig:sddc}
    \includegraphics[width=0.85\linewidth]{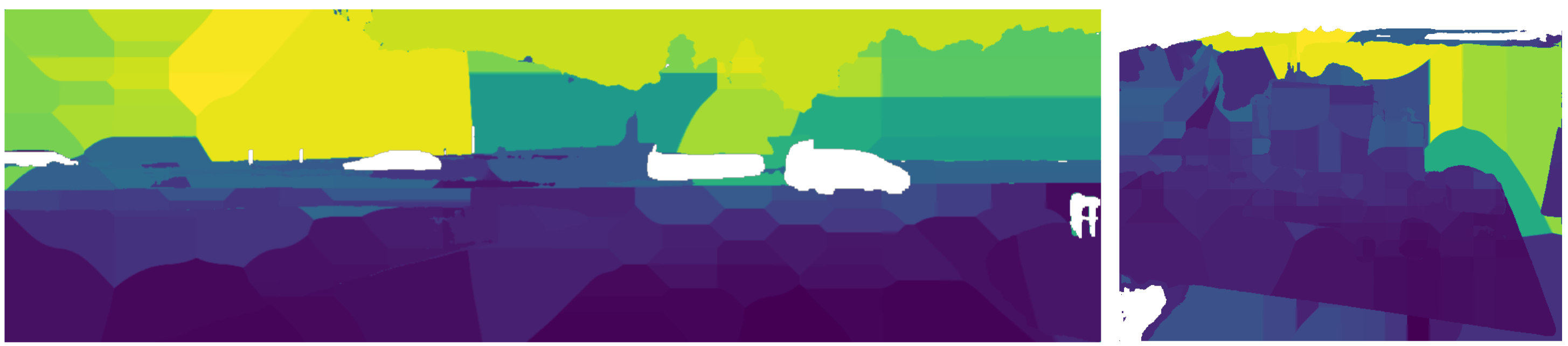}}\vspace{-1mm}
    \caption{Illustration of the sparse depth densification.}
    \label{fig:sdd}
\end{figure}

\subsection{Sparse Depth Densification (SDD)}\label{sec:sdd}

Given the relative pose and the 2D feature correspondences from the pseudo RGB-D SLAM, the sparse depth $D^s$ can be computed using the two-view triangulation~\cite{zhao2020towards}. The triangulation process is formulated as
\begin{equation}
    x^* = \mathop{\arg}\mathop{\max}\limits_{x} \left([d(L_1,x)]^2 + [d(L_2, x)]^2 \right),
\end{equation}
where $L_1$ and $L_2$ denote the camera rays originating from the corresponding 2D features. Here, $d(\cdot)$ measures the geometric distance between a candidate depth point $x$ and the camera rays $L_1$ and $L_2$, and $x$ contains the depth value of each feature within the sparse depth $D^s$.

Upon obtaining the segmentation labels $\mathcal{S}$ from the segmentation network \cite{jain2023oneformer}, we uniformly divide the image into a grid of $d\times d$ rectangular, where $d$ is set to 20 according to the work \cite{liang2024sparse}. Within each local region, the pixels belonging the same category label have similar depth values \cite{liang2024sparse}. In light of this insight, we integrate the category information from $\mathcal{S}$ and depth data from $D^s$ for the densification of sparse depth information.


Specifically, we have the sparse depth points and distinct categories within each grid cell. \figref{fig:sdda} presents examples of gridded images, where the white dots represent the points from $D^s$. Then, when the points are present within the boundaries of the grid cell, the average depth value of these points is propagated throughout the grid region according to the category, as shown in \figref{fig:sddb}.

Furthermore, in the case where the grid region lacks depth points, we conduct a search within the same category to locate the closest five depth points. This search is efficiently performed using the KD-tree algorithm, which can rapidly retrieve the closest depth points. The average depth value of the closest points is then assigned to the entire category within the grid region. The dense depth maps are shown in \figref{fig:sddc}, where the white regions indicate the potential moving objects areas, which are described in \secref{sec:dce}. 

The dense depth map $D^d$ is used to supervise the online learning of the R-DepthNet as pseudo ground-truth. This allows our system to focus more on the target areas with semantic information in the scenarios, such as parked vehicles and buildings, rather than treating all areas equally.

\begin{figure}[tp]
    \centering
    \includegraphics[width=0.4\textwidth]{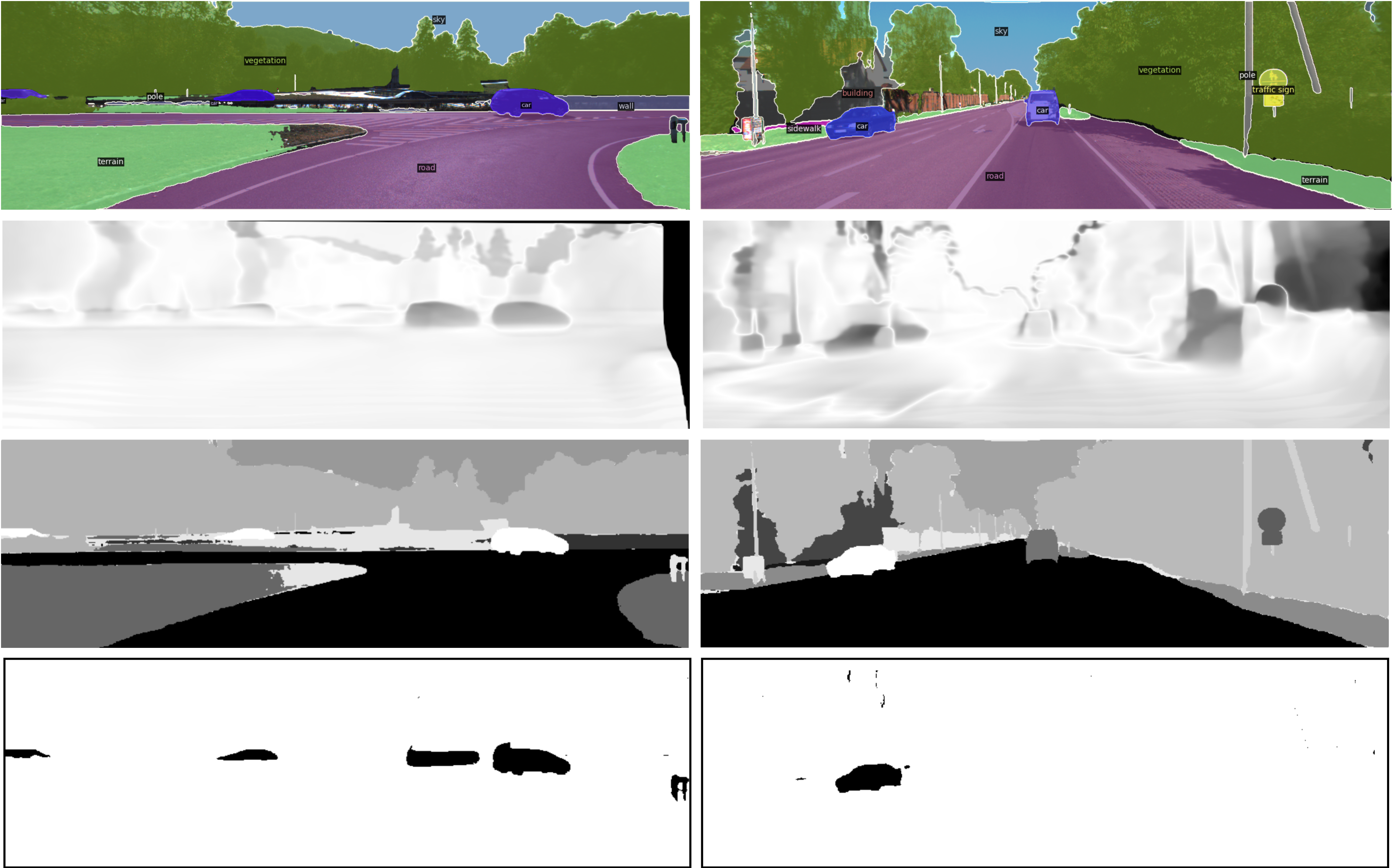}
    \caption{Illustration of two masking processes. \textit{Top to Bottom:} segmented images, self-discovered masks $W_s$ \cite{bian2021unsupervised}, semantic masks $M_{sc}$, and final consistency masks $M$. In $M_{sc}$, the white regions indicate dynamic regions, while the dark regions indicate the regions that conform to the conditions of \equref{equ:M_sc}.}
    \label{fig:dynamic_mask}
\end{figure}

\subsection{Dynamic Consistency Enhancement (DCE)}\label{sec:dce}

In practical applications, moving objects produce non-rigid flow, which violates the geometry consistency assumption and causes the system performance degradation. To address the issue, Bian et al. \cite{bian2021unsupervised} introduce a self-discovered mask $W_s$, which assigns low weights to the geometric inconsistency areas in the predicted depth.
\begin{equation}
    W_s = 1 - |D_{t-1} - D^{t-1}_t| / (D_{t-1} + D^{t-1}_t),
    \label{equ:W_s}
\end{equation}
where $D_{t-1}$ and $D_t$ denote the predicted depth for the image $I_{t-1}$ and $I_t$, respectively, and $D^{t-1}_t$ represents the projected depth of $D_t$ onto the reference frame $D_{t-1}$. However, in the presence of domain shift, the accuracy of the depth estimation network would deteriorate. In this case, it is insufficient to rely solely on the soft mask to assign lower weights to depth inconsistency regions.

In the DCE module, each object within the environments is treated as an integrated entity. We leverage the projection consistency of segmentation masks in adjacent images to handle moving objects. Specifically, we also use the segmentation network~\cite{jain2023oneformer} to segment object regions, which are denoted as $\mathcal{S}_t = \{ s_{t,i}\}_{i=1:n_{t}}$, where $s_{t,i}$ contains the pixel points belonging to the recognized object and $n_t$ is the number of objects in $I_t$. Then, the $s_{t,i}$ is projected into the $I_{t-1}$ space by the relative pose $T_{t,t-1}$ as
\begin{equation}\label{equ:s}
    s_{t,i}^{t-1} = \pi(s_{t,i}, D_t, T_{t,t-1}) = KT_{t,t-1}K^{-1}xD_t[x],
\end{equation}
where $\pi(\cdot)$ is the camera projection function, and $x$ is the pixel point from $s_{t,i}$. The semantic consistency mask $M_{sc}$ can be computed as
\begin{equation}
    M_{sc}(s_{t,i}) = 
    \begin{cases}
        1, &\text{IoU}(s_{t,i}^{t-1}, s_{t-1,j}^*) \ge \tau_s, \\
        0, &{\text{otherwise}},
    \end{cases}
    \label{equ:M_sc}
\end{equation}
where $\text{IoU}(\cdot)$ represents the intersection over union between $s_{t,i}^{t-1}$ and $s_{t-1,j}^*$ as
\begin{equation}
    \text{IoU}(s_{t,i}^{t-1}, s^*_{t-1,j}) = (s_{t,i}^{t-1} \cap s^*_{t-1,j})/(s_{t,i}^{t-1} \cup s^*_{t-1,j}),
\end{equation}
and $s_{t-1,j}^*$ is the object segmentation regions with the highest IoU metric from the $\mathcal{S}_{t-1} = \{ s_{t-1,j}\}_{j=1:n_{t-1}}$ as
\begin{equation}
    s_{t-1,j}^* = \mathop{\arg\max}\limits_{s_{t-1,j} \in \mathcal{S}_{t-1}} \text{IoU}(s_{t,i}^{t-1}, s_{t-1,j}).
\end{equation}
The threshold $\tau_s$ in (\ref{equ:M_sc}) is set to 0.85 in our implementation.
Moreover, following SC-DepthV3 \cite{sun2023sc}, we rank the weights $W_s$ and select the lowest $20\%$ ranks as the potential dynamic regions, which is denoted as $M_{gc}$.

Finally, as illustrated in~\figref{fig:dynamic_mask}, the semantic and geometric consistency mask $M$ can be computed as
\begin{equation}
    M = M_{sc} \cdot M_{gc}.
    \label{equ:M}
\end{equation}

Notably, the mask $M$ is also used to guide the pseudo RGB-D SLAM to extract and select feature points to improve localization performance under dynamic conditions.

\subsection{Loss Functions}\label{sec:lossfunctions}

We use the following loss functions for the online learning of R-DepthNet.

Firstly, we design the depth consistency loss $L_d$ to align the predicted depth $D$ with the densified depth $D_d$ as
\begin{equation}
    L_d = \Vert 1/D - 1/D_d \Vert_1.
\end{equation}

Then, following DepthNet \cite{bian2021unsupervised}, we employ the photometric loss $L_p$ to constrain the warping flow as
\begin{equation}
    L_p = \frac{\alpha}{2}(1- \text{SSIM}(I_t, I_t')) + (1-\alpha) \Vert I_t - I_t'\Vert_1,
\end{equation}
where $I_t'$ denotes the synthesised image from $I_{t-1}$ using the image $I_t$ warping flow. $\text{SSIM}(\cdot)$ is the SSIM loss~\cite{zhou2004image} measuring structural similarity between the two images. Here, $\alpha$ is set to 0.85 according to the work \cite{zhou2004image}. 

Furthermore, we use the geometry consistency loss $L_g$ and the edge-aware loss $L_s$ to ensure the local consistency and smoothness of the the predicted depth
\begin{equation}\label{equ: L_g}
    \begin{aligned}
    L_g &= {\left \| \frac{D_{t-1}' - D_t^{t-1}}{D_{t-1}' + D_t^{t-1}} \right \|_1}, \\
    L_s &= |\partial_x D^*_t| e^{-|\partial_x I_t|} + |\partial_y D^*_t| e^{-|\partial_y I_t|},
    \end{aligned}
\end{equation}
where $D_{t-1}'$ is an interpolation of $D_{t-1}$ to align with $D_t^{t-1}$. $D_t^*$ is the mean-normalized depth. $\partial_x$ and $\partial_y$ are the first derivatives along $x$ and $y$ directions, respectively. 

Finally, the total loss $L_{ol}$ for the online learning is
\begin{equation}\label{eq:L_online}
    L_{ol} = W_s L_p + \lambda_1 L_s+ M(\lambda_2 L_g + \lambda_3 L_d),
\end{equation}
where we set $\lambda_1=\lambda_2=\lambda_3=0.1$ in the implementation.

\section{Experimental Results}
\subsection{Implementation Details}\label{sec:implementationdetails}
The proposed system is implemented using the PyTorch library~\cite{paszke2017automatic} and the Adam optimizer~\cite{Diederik2015adam} with $\beta_1=0.9, \beta_2=0.99$. In the pre-training stage, the DepthNet is trained using the objective function $L_{pre}= W_s L_p + \lambda_1 L_s + \lambda_2 L_g$ for $10^5$ iterations on the DDAD dataset~\cite{guizilini20203d} with a learning rate of $10^{-4}$ and a batch size of $4$. In the online \textcolor{red}{adaptation} stage, we use the objective function $L_{ol}$ with a batch size of $4$. The learning rate is first set to $ 10^{-4}$, and then decays by $0.1$ per $100$ iterations. {We conduct all experiments on a computer with a 3.00GHz Intel i7-9700 CPU and a GeForce RTX 2080 GPU.}

For runtime performance of our system, we adopt a stop learning (SL) strategy for the online \textcolor{red}{adaptation}. Specifically, After $N$ learning steps, if the variance of losses (computed with Exponential Moving Average (EMA)) over $L$ consecutive steps is less than $\tau$, the online learning of R-DepthNet is terminal. The smoothness constant of EMA is set to 0.6. $N$, $L$ and $\tau$ are set to 300, 50, 0.1, respectively.

\subsection{Datasets and Evaluation Metrics}
Our system is pre-trained on the DDAD dataset~\cite{guizilini20203d}. The dataset contains the video sequences of street scenes from different cities. In pre-training, the images are resized to $640 \times 384$ pixels. In the evaluation, we test all the methods on the KITTI~\cite{geiger2012kitti} and TUM~\cite{sturm2012benchmark} dataset, which exhibit significant differences in both environmental contexts and sensing distributions compared to the DDAD dataset. The KITTI dataset captures real-world traffic scenes with many dynamic objects, while the TUM dataset captures indoor scenes with a handheld RGB-D camera. 

For depth estimation evaluation, we first align the scale by computing the ratio of the mean depth values between the predicted depth map and the ground truth\cite{zhan2021df}. Then, we compute the following metrics: absolute relative error (AbsRel), square relative error (SqRel), root mean squared error (RMSE), and the accuracy under threshold ($\delta_i < 1.25^i, i = 1, 2, 3$) as evaluation metrics. For pose estimation evaluation, we use the translation RMSE drift $t_{rel}$ ($\%$) and average rotation RMSE drift $r_{rel}$ (\degree/100m).

\begin{table*}[t]
    \caption{Quantitative pose estimation results on the KITTI}
    \label{tab:kittipose}
\scriptsize
\setlength\tabcolsep{2pt} 
\renewcommand\arraystretch{1.2} 

\resizebox{\textwidth}{!}{
\begin{tabular}{clcccccccccccc}
\toprule \midrule
\multicolumn{2}{c}{\multirow{2}{*}{Method}} & Seq.00 & Seq.01 & Seq.02 & Seq.03 & Seq.04 & Seq.05 & Seq.06 & Seq.07 & Seq.08 & Seq.09 & Seq.10 & Mean \\
\multicolumn{2}{c}{} & $t_{rel}$~~$r_{rel}$ & $t_{rel}$~~$r_{rel}$ & $t_{rel}$~~$r_{rel}$ & $t_{rel}$~~$r_{rel}$ & $t_{rel}$~~$r_{rel}$ & $t_{rel}$~~$r_{rel}$  & $t_{rel}$~~$r_{rel}$  & $t_{rel}$~~$r_{rel}$ & $t_{rel}$~~$r_{rel}$ & $t_{rel}$~~$r_{rel}$  & $t_{rel}$~~$r_{rel}$  & $t_{rel}$~~$r_{rel}$  \\ 
\midrule
 {\multirow{2}{*}{\begin{tabular}[c]{@{}c@{}}Geometry-\\ based\end{tabular}}}
& VINS (S)~\cite{qin2017vins} & 2.91~~0.36  & 59.4~~\textbf{0.21} & 9.82~~0.40 & 2.41~~\textbf{0.19} & 3.00~~{0.18} & \textbf{1.97}~~{0.18} & 2.85~~{0.12}  & 2.15~~0.20 & 4.50~~\textbf{0.18} & 3.91~~0.14  & 2.47~~0.23  & 8.67~~{0.22}  \\
& ORB3 (M)~\cite{campos2021ORBSLAM3} & 14.8~~{0.29}  & 42.3~~{0.64}  & 7.16~~\textbf{0.16} & 2.80~~{0.22} & 2.38~~0.30 & 8.80~~{0.14} & 14.1~~\textbf{0.11}  & 7.40~~{0.12} & 10.4~~0.44 & 9.37~~\textbf{0.13} & 3.94~~{0.13} & 11.2~~{0.21}  \\ \midrule

 {\multirow{5}{*}{\begin{tabular}[c]{@{}c@{}}\textcolor{red}{Self- or Un-} \\supervised\end{tabular}}}
& UnVIO$^\dag$~\cite{wei2021unsupervised} & 3.26~~0.96  & 16.7~~0.61 & 3.11~~0.59 & /~~~~~~/  & 1.95~~0.49 & 3.32~~0.73 & 4.48~~0.92  & 3.49~~0.83 & 4.74~~0.67 & 4.13~~0.89  & 5.51~~0.53  & 5.07~~0.72  \\
& DF-VO$^\dag$~\cite{zhan2021df} & \textbf{2.06}~~0.60 & 42.5~~2.31 & 2.84~~0.74 & 3.35~~0.61 & 1.72~~0.41 & 2.28~~0.70 & \textbf{2.01}~~0.67 & 2.61~~1.55 & 3.95~~0.71 & 3.32~~0.61 & 2.59~~1.06 & 6.29~~0.91 \\
& Bian et al.$^\dag$~\cite{bian2021unsupervised} & 10.0~~3.84 & 25.8~~1.16 & 9.10~~2.16 & 7.52~~2.49 & 3.42~~0.91 & 6.23~~1.78 & 13.5~~2.10 & 6.45~~2.14 & 9.92~~1.98 & 11.5~~3.26 & 10.4~~4.73 & 10.4~~2.41 \\
& BEVO$^\dag$~\cite{chen2023learning} & 3.26~~1.44 & 5.62~~3.95 & 4.56~~1.32 & /~~~~~~/ & 1.79~~1.05 & 2.05~~1.57 & 3.86~~2.08  & 3.58~~1.47 & \textbf{1.94}~~1.04 & \textbf{1.22}~~1.05  & \textbf{1.01}~~1.01  & 2.89~~1.60  \\ 
& \textcolor{red}{Wang et al.$^\dag$ \cite{wang2024unsupervised}} & \textcolor{red}{/~~~~~~/} & \textcolor{red}{/~~~~~~/} & \textcolor{red}{/~~~~~~/} & \textcolor{red}{/~~~~~~/} & \textcolor{red}{/~~~~~~/} & \textcolor{red}{/~~~~~~/} & \textcolor{red}{/~~~~~~/} & \textcolor{red}{/~~~~~~/} & \textcolor{red}{/~~~~~~/} & \textcolor{red}{3.32~~0.41} & \textcolor{red}{3.30~~0.49}  & \textcolor{red}{3.31~~0.45} \\ \midrule

\multirow{4}{*}{Adaptation}
& Xue et al.$^\dag$ \cite{xue2020deep} & /~~~~~~/ & /~~~~~~/ & /~~~~~~/ & 3.32~~2.10 & 2.96~~1.76 & 2.59~~1.25 & 4.93~~1.90 & 3.07~~1.76 & /~~~~~~/ & /~~~~~~/ & 3.94~~1.72 & 3.47~~1.75 \\
& CL-SLAM$^\ast$ \cite{vodisch2022continual} & /~~~~~~/ & /~~~~~~/ & /~~~~~~/ & /~~~~~~/ & 4.37~~0.51 & 4.30~~1.01 & 2.53~~0.63 & \textbf{2.10}~~0.83 & /~~~~~~/ & /~~~~~~/ & 11.2~~1.74  & 4.90~~0.94  \\
& CoVIO$^\ast$ \cite{vodisch2023covio} & /~~~~~~/ & /~~~~~~/ & /~~~~~~/ & /~~~~~~/ & 2.11~~0.53 & 2.88~~0.94 & 2.13~~0.47  & 3.19~~1.26 & /~~~~~~/ & /~~~~~~/ & 3.71~~1.55  & 2.80~~0.95  \\
& Xu et al.$^\ast$ \cite{xu2023online} & /~~~~~~/ & /~~~~~~/ & /~~~~~~/ & /~~~~~~/ & /~~~~~~/ & /~~~~~~/ & /~~~~~~/  & /~~~~~~/ & /~~~~~~/ & 18.1~~6.15 & 26.5~~9.01  & 22.3~~7.76 \\ \midrule

& \textbf{Ours (w/o OL)} & 3.67~~0.32  & 7.74~~0.52  & 3.95~~0.20 & 1.31~~0.22 & 2.43~~0.16 & 3.22~~0.17 & 6.71~~0.32  & 4.14~~0.14 & 3.32~~0.46 & 1.99~~0.45  & 2.81~~{0.18}  & 3.75~~0.29  \\
&  {\textbf{Ours}} & \underline{2.88}~~\textbf{0.30}  & \textbf{5.22}~~\underline{0.31}  & \textbf{2.01}~~\underline{0.19} & \textbf{1.30}~~\underline{0.22} & \textbf{1.42}~~\textbf{0.11} & 2.29~~\textbf{0.14} & 4.66~~\textbf{0.11}  & {3.22}~~\textbf{0.11} & \underline{2.98}~~\underline{0.44} & \underline{1.98}~~\textbf{0.12}  & \underline{1.07}~~\textbf{0.13}  & \textbf{2.64}~~\textbf{0.20}  \\ \midrule


\midrule \bottomrule
\end{tabular}}
\begin{tablenotes}
    \footnotesize
    \item  {$\dag$: pre-training on KITTI, $\ast$: pre-training on Cityscape, and /: no data available.}
\end{tablenotes}

\end{table*}


\begin{table}[t]
    \caption{Quantitative pose estimation results on TUM}
    \label{tab:tumpose}
    \centering
    \small
    \setlength{\tabcolsep}{4pt} 
    \resizebox{0.45\textwidth}{!}{
\begin{tabular}{clccccccc}
\toprule \midrule
\multicolumn{2}{c}{\multirow{2}{*}{Method}} & \multicolumn{4}{c}{Static} & \multicolumn{2}{c}{ {Dynamic}} \\  \cmidrule(r){3-6} \cmidrule(r){7-8} 
 \multicolumn{2}{c}{}  & desk & 360 & slam & long &  {s\_xyz} &  {w\_xyz} \\ \midrule
 {\multirow{2}{*}{\begin{tabular}[c]{@{}c@{}}Geometry-\\ based\end{tabular}} }
 & DSO \cite{engel2017direct} & \ding{55} & \ding{55} & 0.737 & 0.327 &  {0.396} &  {0.736} \\
 & ORB3 (M) \cite{campos2021ORBSLAM3} & 0.115 & \ding{55} & \ding{55} & 0.076 &  {0.434} &  {0.490} \\ \midrule
 {\multirow{3}{*}{\begin{tabular}[c]{@{}c@{}}Self-\\ supervised\end{tabular}}}
 & DF-VO$^\ast$ \cite{zhan2021df} & 0.306 & 0.599 & 0.585 & 0.227 &  {\ding{55}} &  {\ding{55}} \\
 & Zhao et al.$^\ast$ \cite{zhao2020towards} & 0.485 & 0.693 & 0.354 & 0.534 &  {\ding{55}} &  {\ding{55}} \\ 
 & \textcolor{red}{Azzam et al.$^\dag$} \cite{azzam2021a} & \textcolor{red}{/} & \textcolor{red}{/} & \textcolor{red}{0.057} & \textcolor{red}{/} & \textcolor{red}{/} & \textcolor{red}{/} \\
 \midrule
 {\multirow{2}{*}{Adaptation}}
 & Li et al.$^\ast$ \cite{li2021generalizing} & 0.158 & 0.201 & 0.176 & 0.133 &  {/} &  {/} \\
 & Xue et al.$^\dag$ \cite{xue2020deep} & 0.153 & 0.056 & 0.070 & / &  {/} &  {/} \\ \midrule
 & \textbf{Ours (w/o OL)} & 0.119 & 0.056 & 0.370 & 0.070 &  {0.410} &  {0.543} \\
 & \textbf{Ours} & \textbf{0.096} & \textbf{0.031} & \textbf{0.041} & \textbf{0.066} &  {\textbf{0.325}} &  {\textbf{0.428}} \\ 
 \midrule \bottomrule 
\end{tabular}}
\begin{tablenotes}
    \footnotesize
    \item  {$\dag$: pre-training on TUM, $\ast$: pre-training on NYUv2, and \ding{55}: lost.}
\end{tablenotes}
\end{table}

\begin{figure*}[tp]
    \centering
    \includegraphics[width=0.6\linewidth]{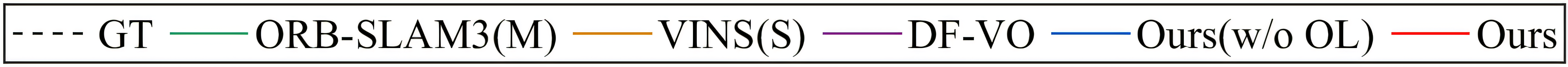} 
    
    \subfigure[KITTI/Seq. 07]{\includegraphics[width=0.3\linewidth]{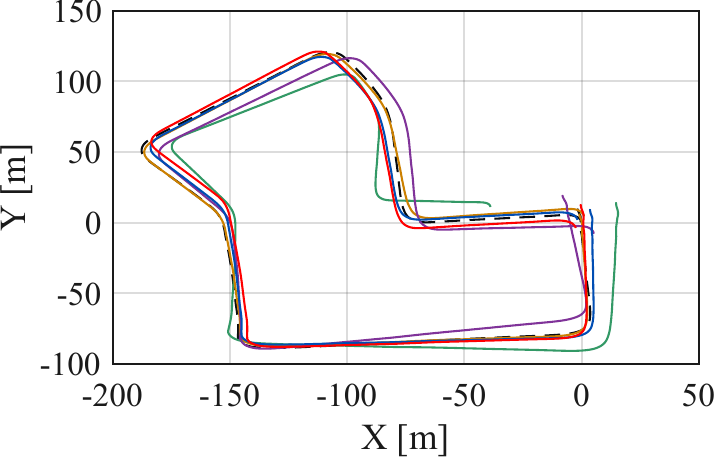}}
    \subfigure[KITTI/Seq. 09]{\includegraphics[width=0.3\linewidth]{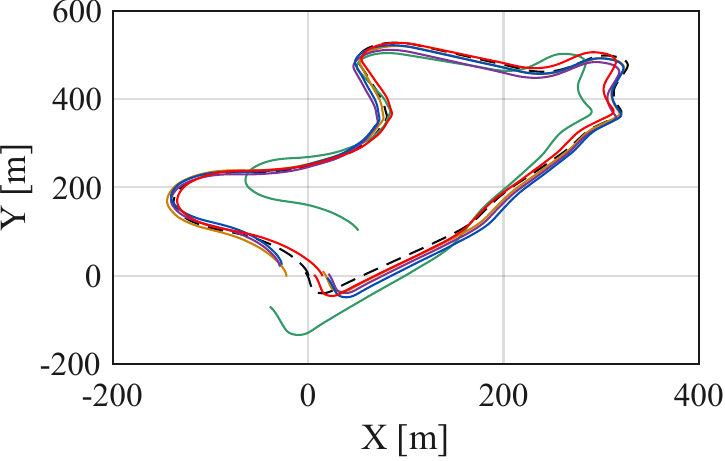}}
    \subfigure[KITTI/Seq. 10]{\includegraphics[width=0.3\linewidth]{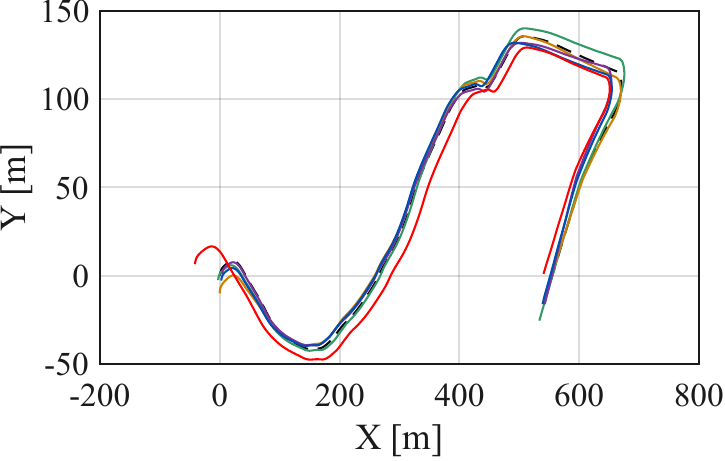}} 
    
    \subfigure[TUM/fr2\_desk]{\includegraphics[width=0.3\linewidth]{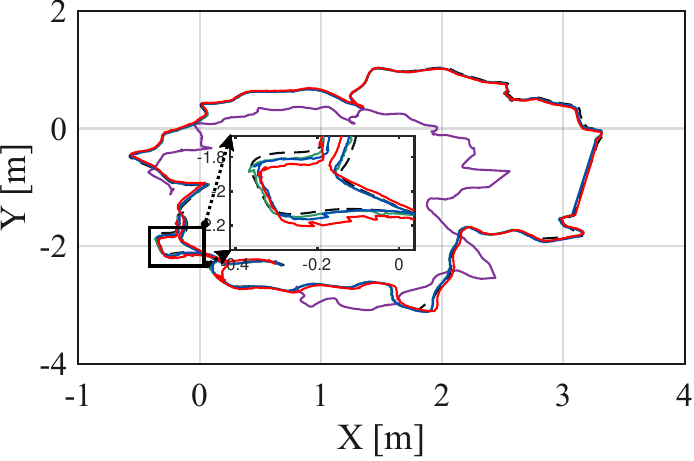}}
    \subfigure[TUM/fr3\_long]{\includegraphics[width=0.3\linewidth]{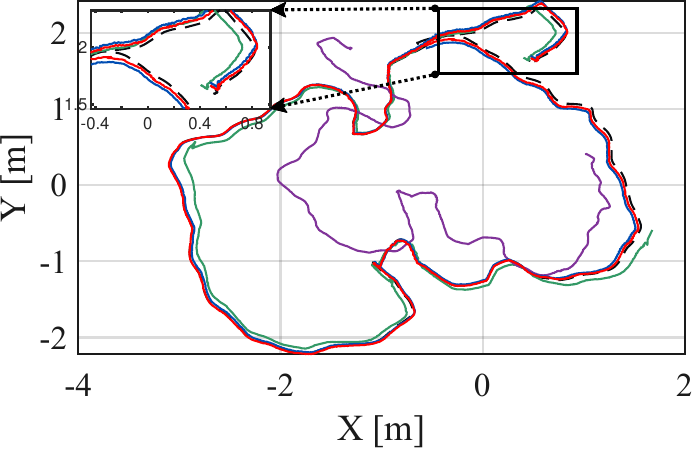}}
    \subfigure[TUM/fr3\_nostr]{\includegraphics[width=0.3\linewidth]{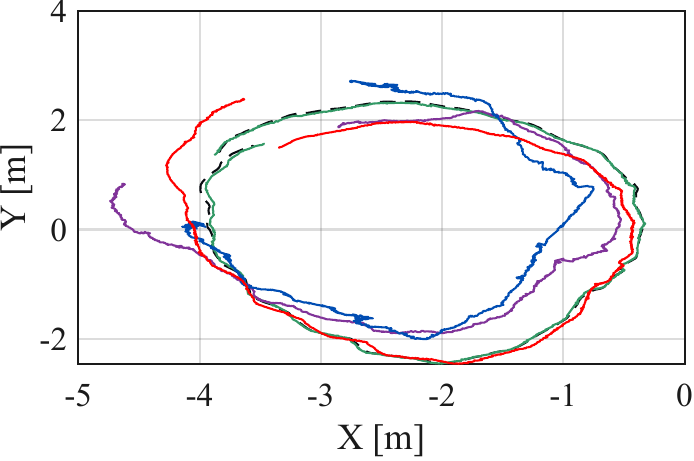}} 
    \caption{Trajectories comparison on the several sequences from the KITTI and TUM datasets.}
    \label{fig:trajs}
\end{figure*}

\subsection{Pose Estimation Evaluation}

In this section, we evaluate our method on pose estimation. \tabref{tab:kittipose} and \figref{fig:trajs} (a)-(c) show visual odometry results on KITTI dataset. We compare with \textit{geometry-based} methods, \textit{self-supervised} methods and \textit{adaptation} methods. `w/o OL' denotes our system without the online learning mechanism. All self-supervised methods are pre-trained on KITTI dataset, while some adaptation methods are pre-trained on Cityscape dataset \cite{cordts2016cityscapes}. It can be seen from \tabref{tab:kittipose}, our method outperforms the self-supervised and adaptation methods on several sequences, and achieves performance comparable to VINS (S).

\tabref{tab:tumpose} and \figref{fig:trajs} (d)-(f) show the visual odometry results {on both static and dynamic sequences from the TUM dataset.} Our method is still pre-trained on the DDAD dataset. The \textit{self-supervised} methods are pre-trained on NYUv2 dataset \cite{silberman2012indoor}. The results demonstrate that the \textit{self-supervised} methods have large errors when encountering the significant domain shift and diverse motion patters. This finding highlights the challenges that these methods face in adapting to new environments. Conversely, our method shows promising results in online learning, and more accurate than the \textit{online adaptation} methods and more robust than the \textit{geometry-based} methods.

We also develop the proposed method to a mobile robot in the outdoor scene.\footnote{Our collected data is available at \url{https://drive.google.com/drive/folders/1_7EOaonetNQCJxYQ2yDf9EVMCcMv6qeh?usp=sharing}} The mobile robot is illustrated in~\figref{fig:robot}, and the trajectory comparison is illustrated in~\figref{fig:rw}.

\begin{figure}[t]
    \centering
    \includegraphics[width=0.6\linewidth]{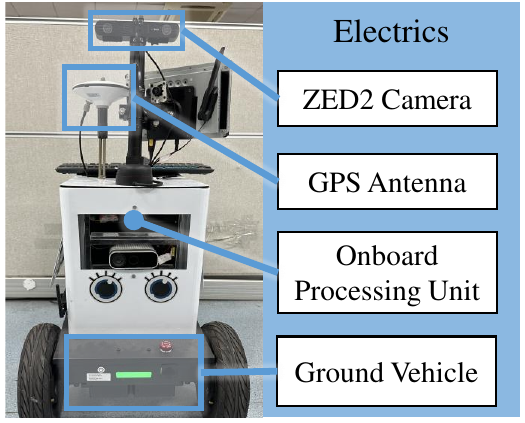}
    \caption{The mobile robot platform.}
    \label{fig:robot}
\end{figure}

\begin{figure}[t]
    \centering
    \includegraphics[width=0.85\linewidth]{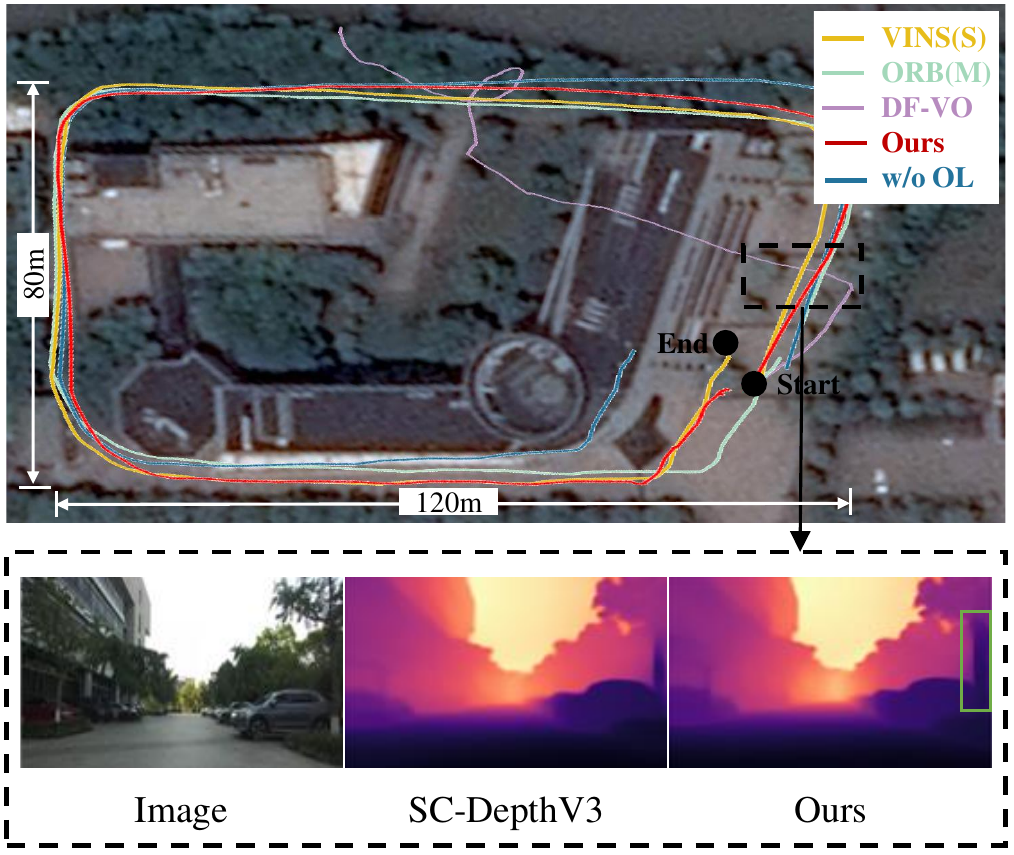}
    \caption{Comparison of trajectories and estimated depth in the outdoor scene.}
    \label{fig:rw}
\end{figure}

\begin{table}[t]
    \caption{Quantitative depth estimation results on KITTI}
    \label{tab:kittidepth}
    \centering
    \small
\resizebox{0.48\textwidth}{!}{
\begin{tabular}{clcccccc}
    \toprule \midrule
    \multirow{2}{*}{ {Source}} & \multirow{2}{*}{Methods} & \multicolumn{3}{c}{Errors} & \multicolumn{3}{c}{Accuracy} \\ \cmidrule(r){3-5} \cmidrule(r){6-8} 
     &  & AbsRel & SqRel & RMSE & $\delta_1$ & $\delta_2$ & $\delta_3$ \\ \midrule
    \multirow{4}{*}{KITTI} & SfMLearner~\cite{zhou2017unsupervised} & 0.327 & 3.113 & 9.522 & 0.423 & 0.701 & 0.848 \\
     & IMUSAtt~\cite{zhang2024selfsupervised} & 0.115 & 0.640 & 4.106 & 0.862 & 0.960 & 0.986 \\
     & Bian et al.~\cite{bian2021unsupervised} & 0.163 & 0.940 & 4.913 & 0.776 & 0.932 & 0.977 \\
     & SC-DepthV3~\cite{sun2023sc} & \textbf{0.092} & \textbf{0.442} & \textbf{3.698} & \textbf{0.916} & \textbf{0.984} & \textbf{0.990} \\ \midrule
    \multirow{4}{*}{DDAD} 
     & Bian et al.~\cite{bian2021unsupervised} & 0.223 & 2.130 & 6.373 & 0.701 & 0.898 & 0.959 \\
     & \textbf{with Ours} & \textbf{0.168} & \textbf{1.062} & \textbf{5.070} & \textbf{0.764} & \textbf{0.932} & \textbf{0.977} \\ \cmidrule(r){2-8} 
     & SC-DepthV3~\cite{sun2023sc} & 0.254 & 2.910 & 7.449 & 0.653 & 0.874 & 0.953 \\
     & \textbf{with Ours} & \textbf{0.196} & \textbf{1.227} & \textbf{5.645} & \textbf{0.663} & \textbf{0.902} & \textbf{0.973} \\ 
     \midrule \bottomrule
\end{tabular}}
\end{table}


\begin{table}[t]
    \caption{Quantitative depth estimation results on TUM}
    \label{tab:tumdepth}
    \centering
    \small
\resizebox{0.48\textwidth}{!}{
\begin{tabular}{clcccccc}
\toprule \midrule
\multirow{2}{*}{ {Source}} & \multirow{2}{*}{Methods} & \multicolumn{2}{c}{Static} & \multicolumn{2}{c}{ {Dynamic}} & \multicolumn{2}{c}{ {Mean}} \\  \cmidrule(r){3-4} \cmidrule(r){5-6} \cmidrule(r){7-8} 
 &  & AbsRel & $\delta_1$ & AbsRel & $\delta_1$ & AbsRel & $\delta_1$ \\ \midrule
 
\multirow{3}{*}{TUM} & MonoDepth2 \cite{godard2019digging} & 0.312 & 0.502 &  {0.328} &  {0.496} &  {0.317}&  {0.500} \\
 & Bian et al. \cite{bian2021unsupervised} & 0.198 & 0.667 &  {0.220} &  {0.722} &  {0.205} &  {0.685} \\
 & SC-DepthV3 \cite{sun2023sc} & \textbf{0.182} & \textbf{0.731} &  {\textbf{0.153}} &  {\textbf{0.800}} &  {\textbf{0.172}} &  {\textbf{0.754}} \\ \midrule

 
\multirow{4}{*}{DDAD} 
& Bian et al. \cite{bian2021unsupervised} & 0.387 & 0.456 &  {0.423} &  {0.515} &  {0.399} &  {0.476} \\
& \textbf{with Ours} & \textbf{0.308} & \textbf{0.676} &  {\textbf{0.354}} &  {\textbf{0.565}} &  {\textbf{0.323}} &  {\textbf{0.639}} \\ \cmidrule(r){2-8} 
& SC-DepthV3 \cite{sun2023sc} & 0.389 & 0.495 &  {0.485} &  {0.374} &  {0.421} &  {0.455} \\
& \textbf{with Ours} & \textbf{0.299} & \textbf{0.525} &  {\textbf{0.361}} &  {\textbf{0.467}} &  {\textbf{0.320}} &  {\textbf{0.506}} \\ 
\midrule \bottomrule
\end{tabular}}
\end{table}

\begin{table}[!t]
    \caption{Comparison of methods pre-trained on diverse datasets}
    \label{tab:tumdepth_multisource}
    \centering
    \small
    \setlength\tabcolsep{2pt} 
    \renewcommand\arraystretch{1} 
\resizebox{0.47\textwidth}{!}{
\begin{tabular}{clcccccc}
\toprule \midrule
\multirow{2}{*}{Source} & \multirow{2}{*}{Methods} & \multicolumn{2}{c}{desk} & \multicolumn{2}{c}{360} & \multicolumn{2}{c}{long} \\  \cmidrule(r){3-4} \cmidrule(r){5-6} \cmidrule(r){7-8} 
 &  & AbsRel & $\delta_1$ & AbsRel & $\delta_1$ & AbsRel & $\delta_1$ \\ \midrule
\multirow{2}{*}{8 datasets}
& DepthAnythingV2 \cite{yang2024depth} 
& 0.200 & 0.787 & 0.173 & 0.703 & 0.204 & 0.764 \\ 
& \textbf{with Ours} 
& \textbf{0.185} & \textbf{0.839} & \textbf{0.125} & \textbf{0.822} & \textbf{0.177} & \textbf{0.780} \\ \midrule

 
\multirow{2}{*}{2 datasets}
& Marigold \cite{ke2024repurposing} 
& 0.301 & 0.537 & 0.362 & 0.414 & 0.352 & 0.437 \\
& \textbf{with Ours} 
& \textbf{0.235} & \textbf{0.688} & \textbf{0.329} & \textbf{0.443} & \textbf{0.226} & \textbf{0.625} \\ 
\midrule \bottomrule

\end{tabular}}
\end{table}

\subsection{Depth Estimation Evaluation}
In the section, we evaluate our method on depth prediction. \tabref{tab:kittidepth} shows the comparison with the state-of-the-art approaches on KITTI dataset. Following \cite{ji2022georefine}, we project the laser point cloud onto the image plane of the left camera to generate ground-truth depths.
SfMLeaner~\cite{zhou2017unsupervised}, Bian et al.~\cite{bian2021unsupervised} and IMUSAtt~\cite{zhang2024selfsupervised} are unsupervised-based methods, where IMUSAtt incorporates IMU data to enhance depth estimation accuracy. SC-DepthV3~\cite{sun2023sc} employs an additional supervision module to refine the depth maps. These methods are all pre-trained on the KITTI dataset. Besides, we pre-train Bian et al.~\cite{bian2021unsupervised} and SC-DepthV3~\cite{sun2023sc} on the DDAD dataset and integrate our method into them. The results demonstrate that our method significantly improves the performance of the two methods in adapting to the KITTI dataset. Furthermore, the work of Bian et al., when integrated with our method, results in performance close to that of pre-training on the KITTI dataset. 

\tabref{tab:tumdepth} presents the results obtained on TUM dataset, where the depth maps captured by the Kinect camera are treated as ground-truth. \tabref{tab:tumdepth_multisource} compares the performance of methods pre-trained on various datasets (excluding TUM).  \figref{fig:depth} provides a visual comparison of the estimated depth on KITTI and TUM dataset, illustrating the qualitative improvements in depth estimation before and after online learning. Additionally, \figref{fig:rw} shows a visual comparison in the outdoor scene. These results collectively indicate that our method generalizes effectively across diverse approaches and datasets.

\textcolor{red}{The uncertainty of densified depths heavily depends on the quality of semantic segmentation and sparse depth points. In cases where these inputs are insufficient or incorrect, the densified depths exhibit significant errors, as illustrated in \figref{fig:failure}.} However, such failure cases are infrequent, and our method generally performs well on the TUM and KITTI datasets, \textcolor{red}{underscoring its ability to handle uncertainty in most scenarios.}


\begin{figure}[tp]
    \centering
    \subfigure[KITTI]{\includegraphics[width=0.95\linewidth]{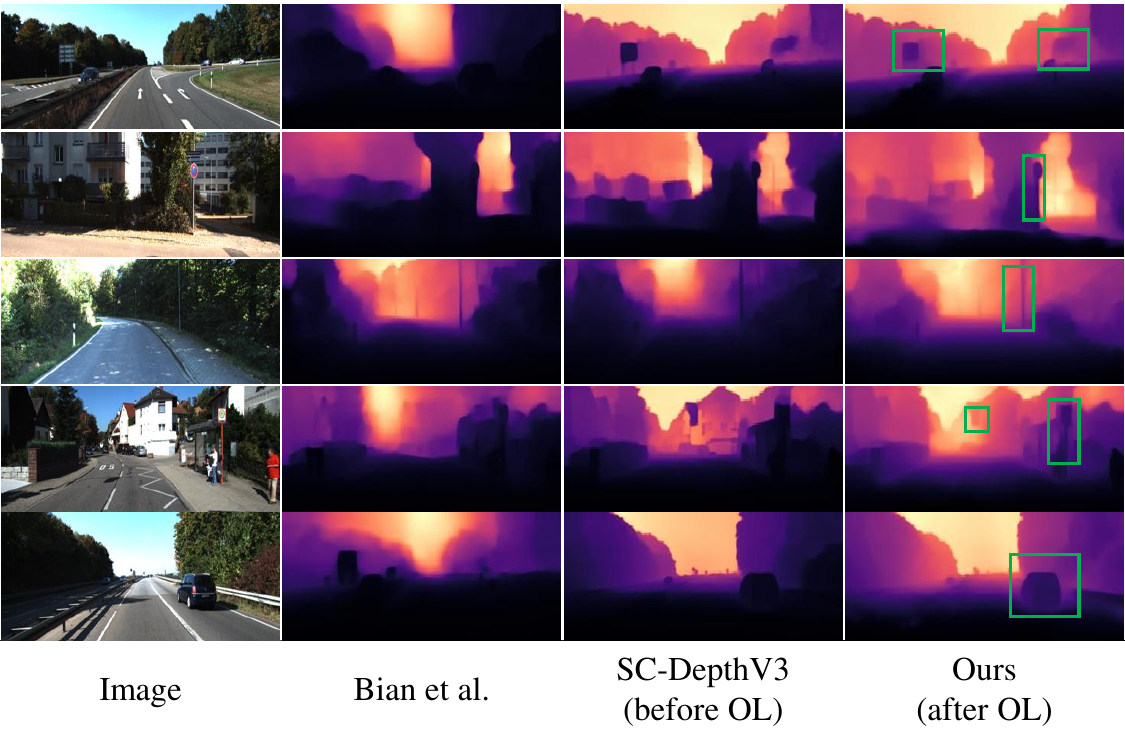}\label{fig:kitti_depth}} 
    
    \subfigure[TUM]{\includegraphics[width=0.95\linewidth]{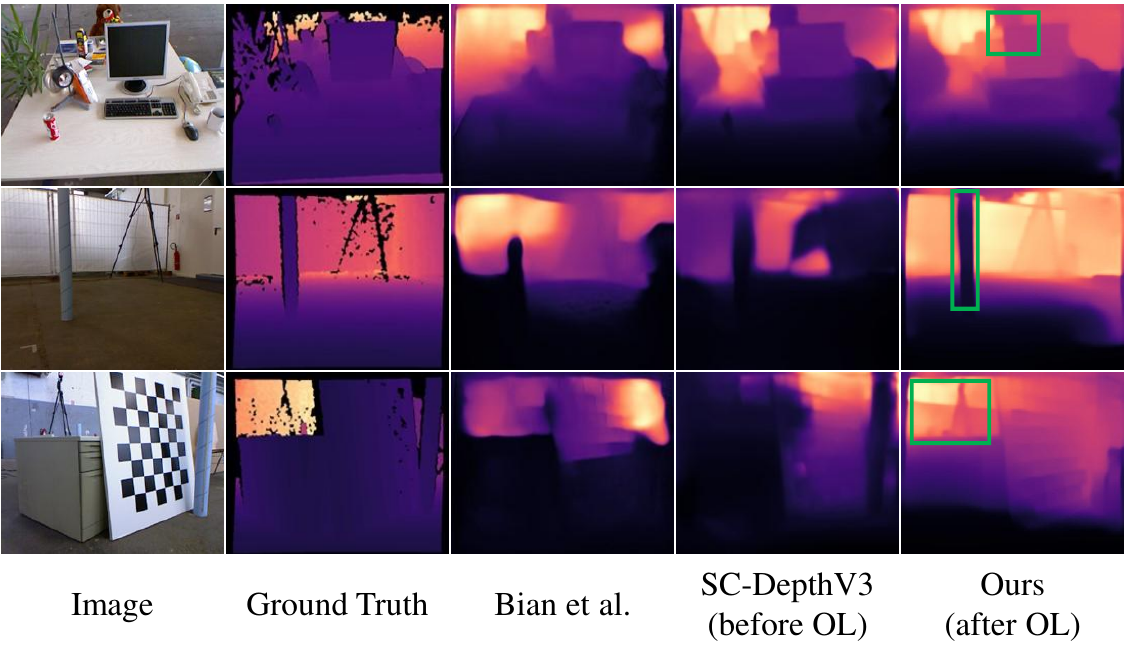}\label{fig:tum_depth}} 
    \caption{Visual comparison of estimated depth on the KITTI and TUM dataset. Regions with salient improvement are highlighted with green boxes.}  
    \label{fig:depth}
\end{figure}

\begin{figure}[t]
    \centering
    \subfigure[Sensitive to Segmentation]{\includegraphics[width=0.45\linewidth]{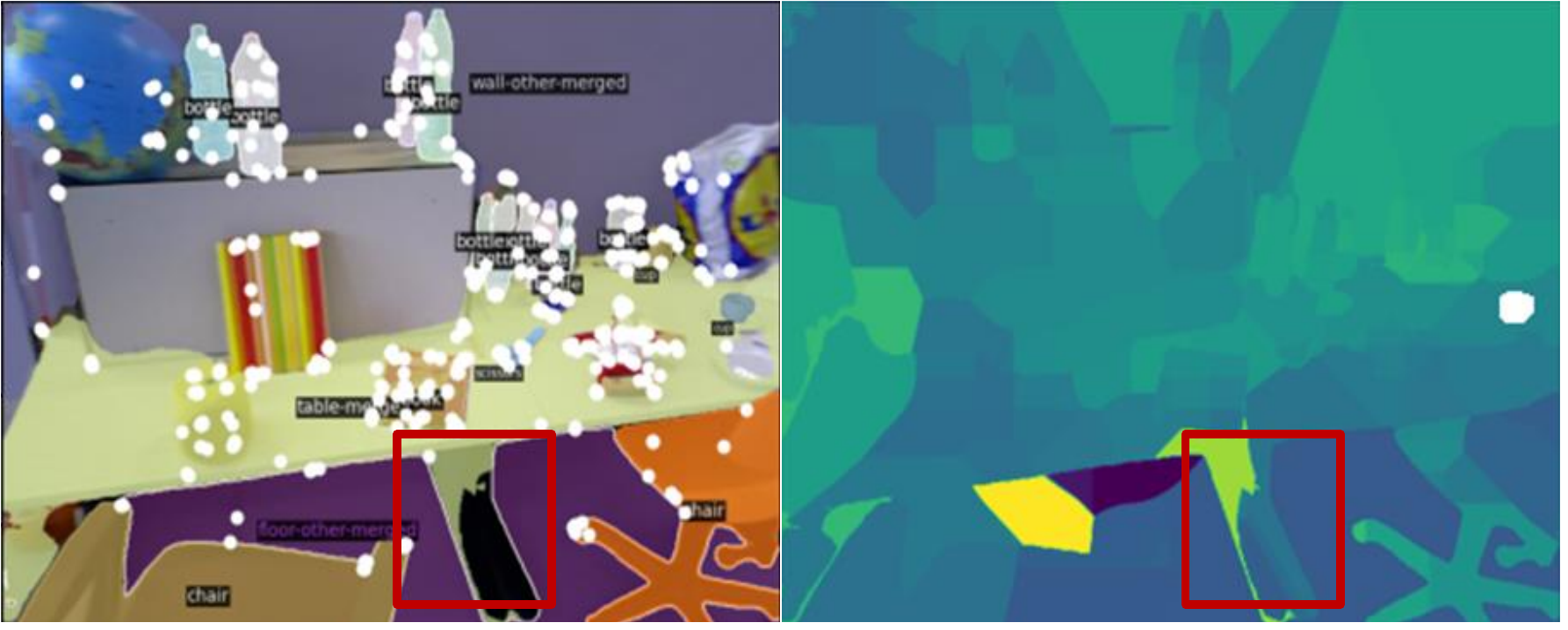}\label{fig:example2}}
    \subfigure[Depth Measurements Issue]{\includegraphics[width=0.46\linewidth]{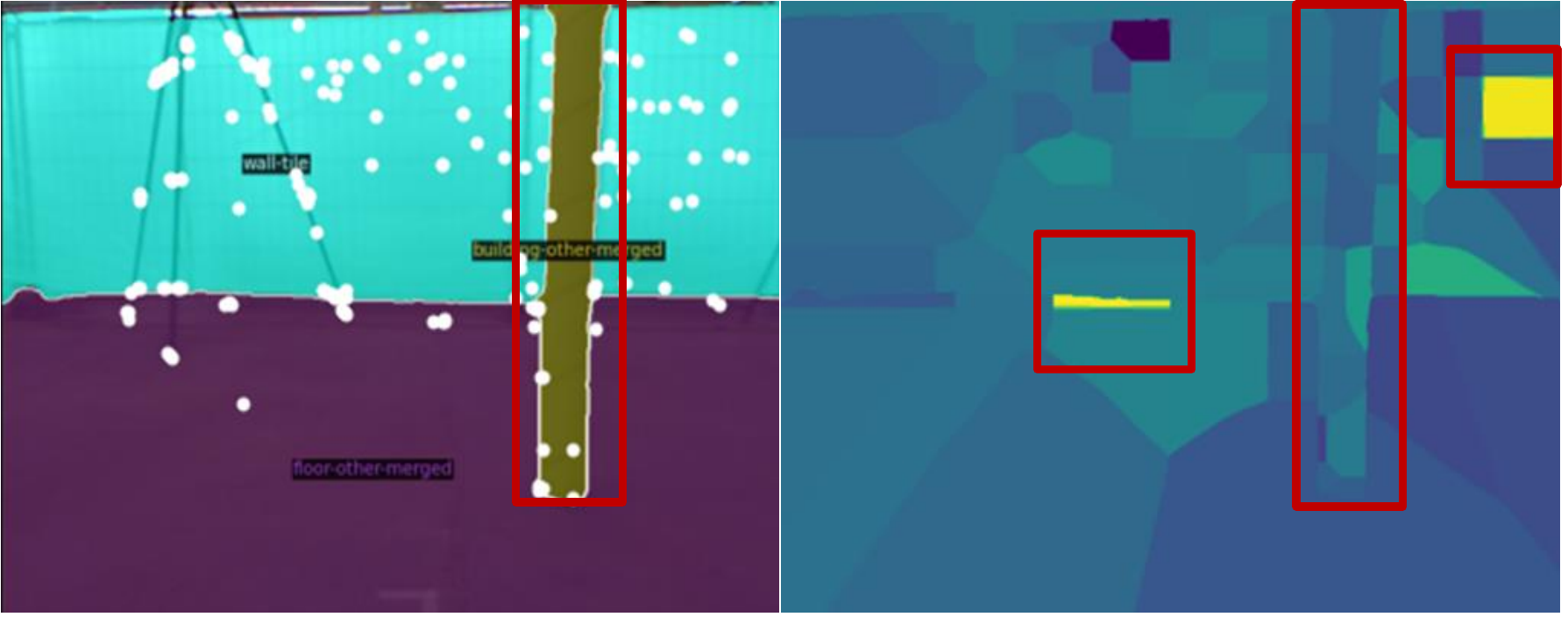}\label{fig:example1}}
    \caption{{Examples of densified depth in the cases.}}
    \label{fig:failure}
\end{figure}

\begin{figure}[tp]
    \centering
    \includegraphics[width=0.8\linewidth]{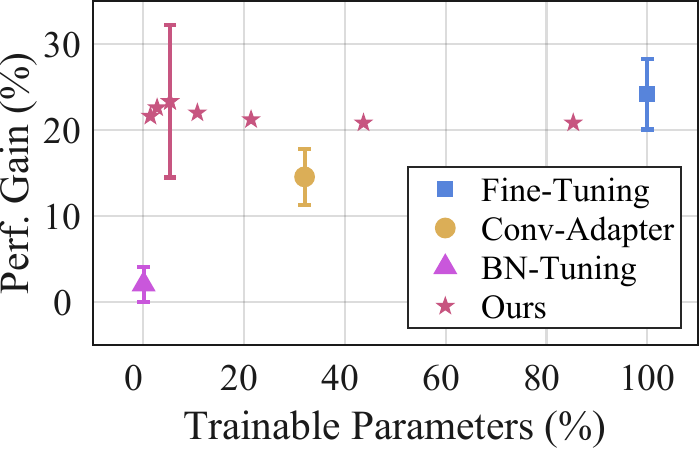}
    \caption{Evaluation of the our refiner module on KITTI and TUM datasets.}
    \label{fig:AAM}
\end{figure}

\begin{figure}[tp]
    \centering
    \includegraphics[width=0.8\linewidth]{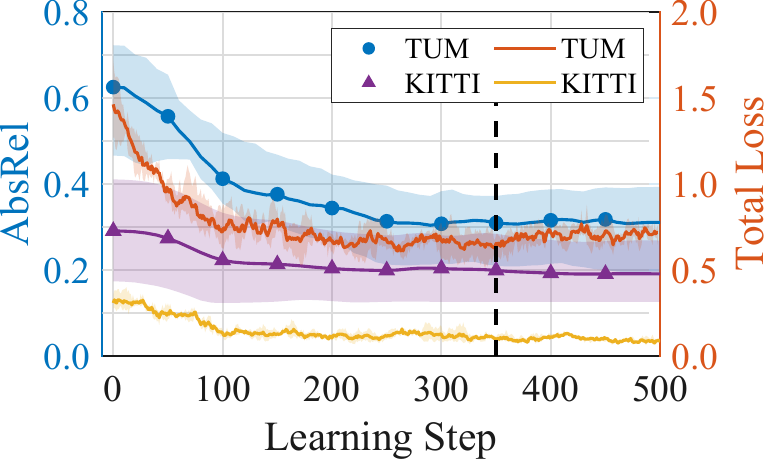}
    \caption{AbsRel and total loss versus the learning step.}
    \label{fig:TLS}
\end{figure}

\begin{table}[t] 
    \caption{Comparison of the learning methods on TUM}
    \label{tab:runtime}
    \centering
    \small
    \setlength\tabcolsep{4pt} 
    \renewcommand\arraystretch{1} 
    \resizebox{0.45\textwidth}{!}{
\begin{tabular}{lcccc}
\toprule \midrule
                                                            & \begin{tabular}[c]{@{}c@{}}Fine-Tuning\\  \cite{li2021generalizing}\end{tabular}
                                                            & \begin{tabular}[c]{@{}c@{}}BN-Tuning\\ \cite{wang2021tent}\end{tabular}
                                                            & \begin{tabular}[c]{@{}c@{}}Conv-Adapter\\ \cite{chen2022conv}\end{tabular}
                                                            & Ours  \\ \midrule
Runtime         & 533ms & 308ms  & 338ms  & 432ms   \\ 
\midrule \bottomrule
\end{tabular}}
\end{table} %




\subsection{Self-supervised Online Learning} \label{sec:theRefinerModule}
We compare our refiner module with several transfer learning methods, including fine-tuning pre-trained weights (Fine-Tuning) \cite{li2021generalizing}, fine-tuning batch norm layers weights (BN-Tuning) \cite{wang2021tent} and the conv-adapter method (Conv-Adapter) \cite{chen2022conv}. As shown in \figref{fig:AAM}, our method not only maintains the number of trainable parameters as BN-Tuning, but also achieves comparable performance to Fine-Tuning. {`Perf. Gain' refers to the performance gain relative to the baseline, i.e. the reduction ratio of the AbsRel.} The symbols `Ours' from left to right indicate the parameter $r$ takes values in the range $\{2,4,8,16,32,128 \}$, where $r=8$ achieves optimal performance.

{\tabref{tab:runtime} presents a comparison of these learning methods. `Runtime' refers to average runtime for a single learning step. Typically, our refiner model converges within 200 steps, while finetuning the entire network needs more than 300 steps to converge.}
{\figref{fig:TLS} shows AbsRel and total loss versus the learning step. The online learning terminates at the vertical dashed line.}

\begin{table}[!ht]
    \caption{Ablation study on KITTI Seq. 09}
    \label{tab:ablationstudy}
    \centering
    \small
    \resizebox{0.45\textwidth}{!}{
\begin{tabular}{lcccccc}
\toprule \midrule
\multirow{2}{*}{Methods} & \multicolumn{3}{c}{Errors} & \multicolumn{3}{c}{Accuracy} \\ \cmidrule(r){2-4} \cmidrule(r){5-7} 
 & AbsRel & SqRel & RMSE & $\delta_1$ & $\delta_2$ & $\delta_3$ \\ \midrule
Baseline~\cite{sun2023sc} & 0.228 & 3.298 & 9.265 & 0.701 & 0.882 & 0.956 \\
Ours & \textbf{0.169} & \textbf{1.014} & \textbf{5.420} & \textbf{0.755} & \textbf{0.920} & \textbf{0.973} \\ \midrule
\multicolumn{7}{c}{\textbf{Ablation on each component}} \\ \midrule
w/o SDD & 0.178 & 1.469 & 6.466 & 0.727 & 0.914 & 0.973 \\ 
w/o DCE & 0.181 & 1.306 & 6.680 & 0.719 & 0.915 & 0.970 \\ \midrule
\multicolumn{7}{c}{\textbf{Ablation on masks of the DCE module}} \\ \midrule
w/o $M_{sc}$ & 0.182 & 1.109 & 5.987 & 0.709 & 0.906 & 0.966 \\
w/o $M_{gc}$ & 0.186 & 1.133 & 5.968 & 0.699 & 0.893 & 0.959 \\ \midrule
\multicolumn{7}{c}{ {\textbf{Ablation on densified depth maps}}} \\ \midrule

 {with SDM} &  {0.201} &  {1.266} &  {6.504} &  {0.684} &  {0.905} &  {0.969} \\
 {with BP-Net \cite{tang2024bilateral}} &  {0.177} &  {1.261} &  {6.399} &  {0.715} &  {0.907} &  {0.965} \\

\midrule \bottomrule
\end{tabular}}
\end{table}

\subsection{Ablation Study}
\tabref{tab:kittipose}, \tabref{tab:tumpose}, \tabref{tab:kittidepth} and \tabref{tab:tumdepth} show results with and without our online \textcolor{red}{adaptation} method about pose estimation and depth prediction, which have demonstrated the efficiency of the online learning. We conduct further ablation studies on depth estimation. Note that we fix the pseudo RGB-D SLAM to avoid any impact on depth estimation.

\tabref{tab:ablationstudy} provides the results of ablation study on each component and masks of the DCE module. The results demonstrate the effectiveness of the SDD and DCE module, and the importance of incorporating both semantic and geometric consistency masks. 

In the ablation study on the densified depth maps, `with SDM' refers to directly using sparse depth maps instead of densified depth maps. `with BP-Net' refers to using sparse-to-dense depth maps from BP-Net \cite{tang2024bilateral} instead of densified depth maps.
BP-Net predicts dense depths from sparse depth measurements of a LiDAR sensor with synchronized color images. We input the sparse depth maps obtained from the SLAM and RGB images into BP-Net.

\begin{figure}[tp]
    \centering
        \subfigure[]{
    \includegraphics[width=0.47\linewidth]{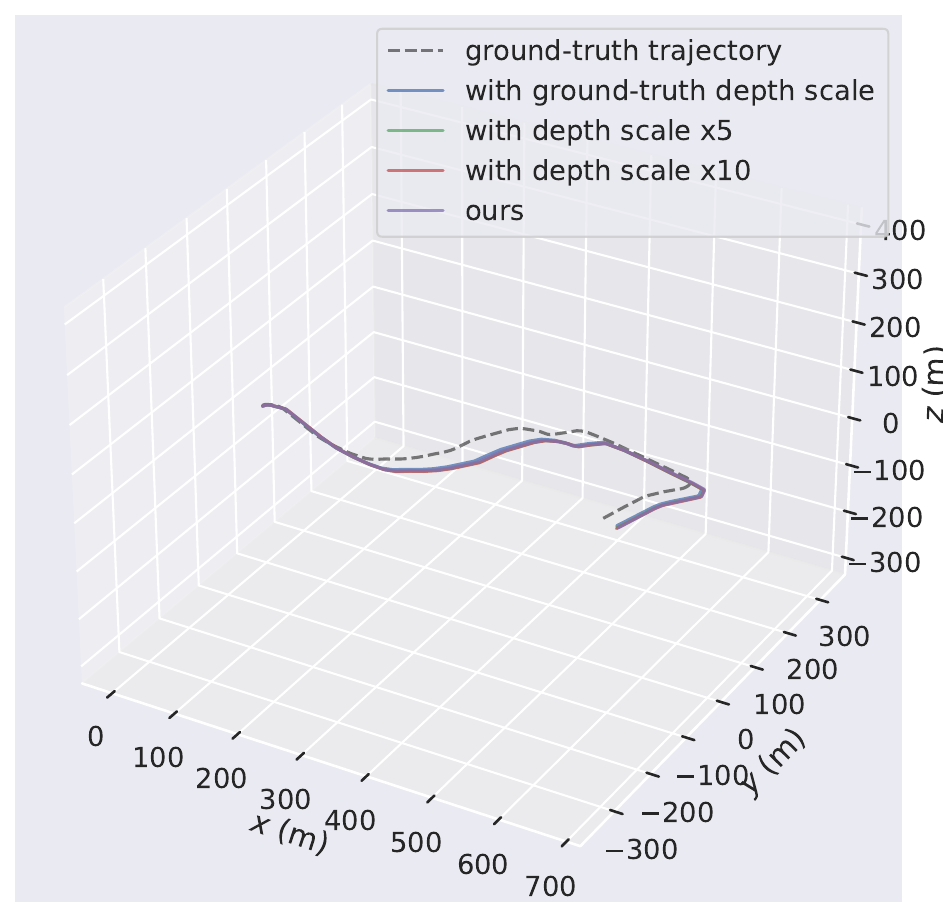}\label{fig:traj_odom10}}
    \subfigure[]{
    \includegraphics[width=0.47\linewidth]{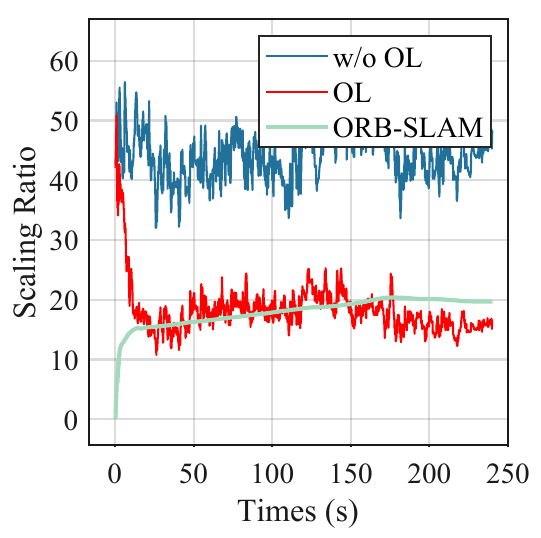}\label{fig:scale_odom10}}

    \caption{(a) Comparison of trajectories with different scales of pseudo depth maps. (b) Scaling ratio (w.r.t ground-truth) versus times.}
\end{figure}

We further provide ablation experimental studies on the scale consistency. We consider three cases in our online \textcolor{red}{adaptation}:
\begin{enumerate}
    \item The scale of the depth map of R-DepthNet is aligned with the ground-truth, denoted by `with ground-truth depth scale'.
    \item The scale of the depth map of R-DepthNet is arbitrarily scaled by 5 or 10 times, denoted by `with depth scale x5' and `with depth scale x10', respectively.
    \item No modification on the scale of the depth map of R-DepthNet, denoted by `ours'.
\end{enumerate}
\figref{fig:traj_odom10} compares the result trajectories in these cases on two data sequences from KITTI odom 10. It can be seen that the scale has little effect on the results.

\begin{figure}[t]
    \centering
    \subfigure[KITTI]{\includegraphics[width=0.9\linewidth]{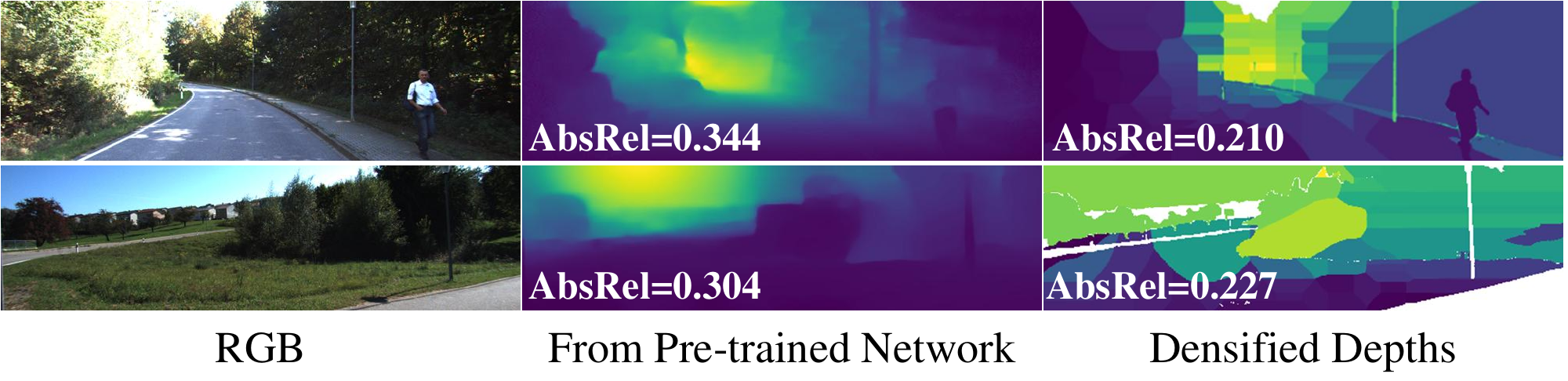}}
    
    \subfigure[TUM]{\includegraphics[width=0.9\linewidth]{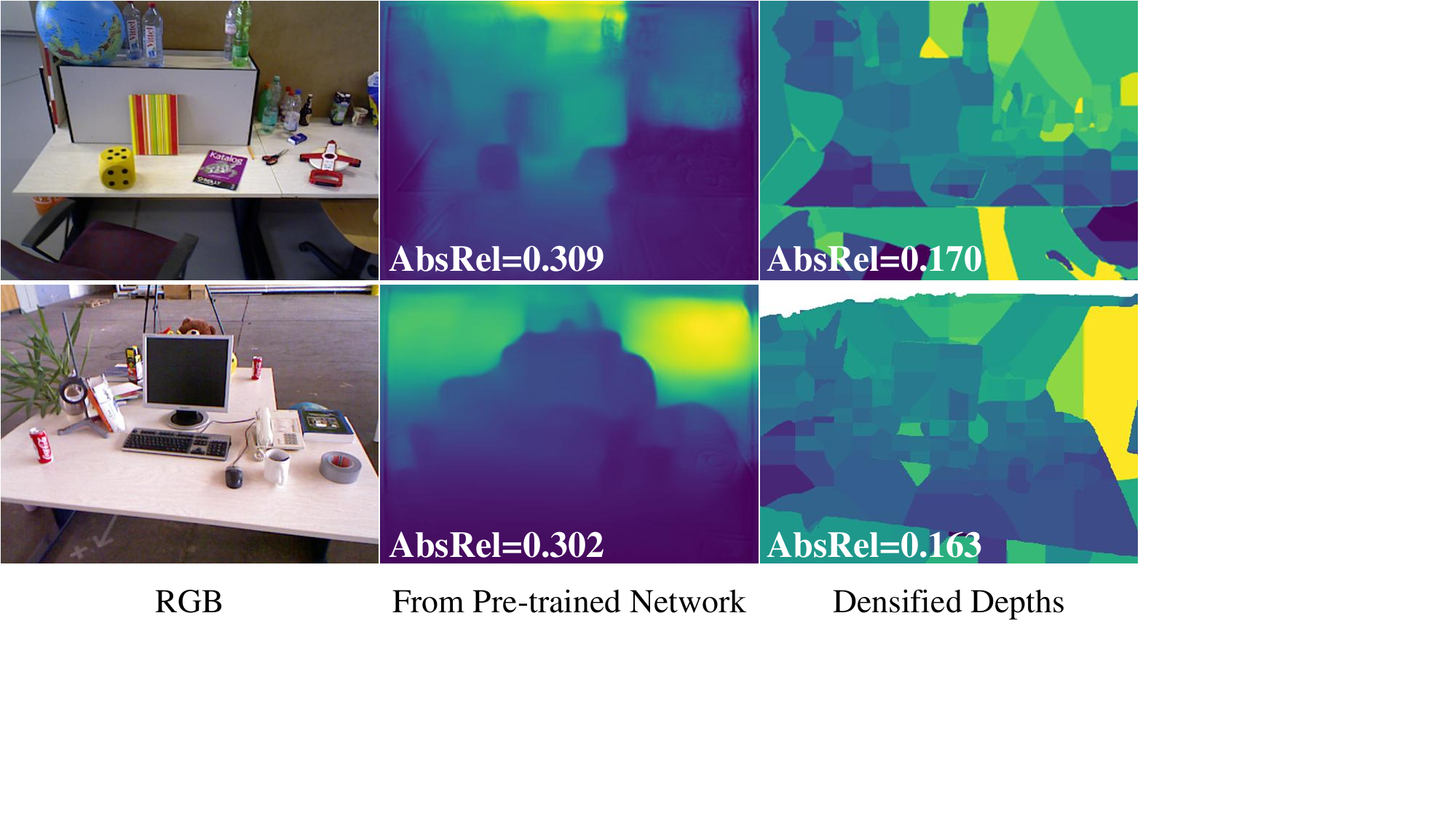}}
    \caption{{Comparisons between densified depths and the depths from pre-trained network \cite{sun2023sc}.}}
    \label{fig: comparion with pretrained network}
\end{figure}

\section{Discussion}
In this section, we discuss why locally coarse depth densification is effective for network \textcolor{red}{adaptation}, and how the scale of pseudo depth \textcolor{red}{measurements} affects pose estimation.


\subsection{Effectiveness of Depth Densification}
The SDD module simplifies depth construction by assuming that pixels within the same grid sharing identical semantic labels also share the same depth \textcolor{red}{measurement}. While this assumption may \textcolor{red}{introduce potential uncertainty, particularly in locally coarse densified depth regions}, its overall quality can be better than the depth estimation by the network pre-trained on the source domain. As illustrated in \figref{fig: comparion with pretrained network}, the AbsRel of the densified depth map can be lower than that of the estimation by the source pre-trained network. Our ablation studies demonstrate that using densified depth map yields significantly better performance than sparse depth map (`with SDM'). Furthermore, under this assumption, the densified depth maps are approximately unbiased in statistics. Meanwhile, our method freezes pre-trained network, and only finetunes the lightweight adapter modules. The proportion of trainable parameters is only $5\%$, which can prevent our method overfitting to erroneous depths. Therefore, the densified depth map is effective to supervise the online learning of the pre-trained networks \cite{bian2021unsupervised, sun2023sc, yang2024depth, ke2024repurposing}.

\subsection{Effect of depth scale}
Typically, the following residual $e_s$ is used for RGB-D SLAM \cite{mur2017ORBSLAM2}.


\begin{equation}\label{equ: e_s}
    \begin{aligned}
        e_s = 
        \begin{bmatrix}
        p_{uv} \\
        u_r
        \end{bmatrix}
        - 
        \begin{bmatrix}
        f_x \frac{x}{z} + c_x \\
        f_y \frac{y}{z} + c_y \\
        f_x \frac{sx - b}{sz} + c_x
        \end{bmatrix},
    \end{aligned}
\end{equation}
where $b$ is the baseline. We use the residual $e$ in \equref{equ: e} instead of $e_s$, because R-DepthNet’s scale performance is unstable when adapting to new environments. \figref{fig:scale_odom10} shows the scales of pre-trained R-DepthNet on source domain (`w/o OL'), adapted R-DepthNet via online learning (`OL'), and ORB-SLAM \cite{campos2021ORBSLAM3} on KITTI odom 10. Although the scale of R-DepthNet in online learning gradually converges to be consistent with that of ORB-SLAM, the scale varies greatly between adjacent frames. It can make the optimization based on the residual $e_s$ converge slowly or fail. In constrast,
the scale factor $s$ in \equref{equ: e} is eliminated during the optimization process. \figref{fig:traj_odom10} illustrates when the pseudo depth maps with various scale are used as input, the resulting trajectories are nearly identical. Therefore, our proposed system is insensitive in the scale of the pseudo depth map during the online learning.

\section{Conclusion}
In this paper, we present a {self-supervised} online \textcolor{red}{adaptation} framework of \textcolor{red}{depth estimation and visual odometry} for fast adaptation to novel environments, in which the depth estimation module and the pseudo RGB-D SLAM reinforce each other and improve their respective performance in the online manner. Extensive experiments across various datasets and on the mobile robot demonstrate the efficiency and adaptation capability of our system to novel environments. In the future, we will further explore active learning methods for autonomous navigation systems to improve the flexibility of robots in unknown environments.

\bibliographystyle{IEEEtran}
\bibliography{Bibliography/reference}

\begin{IEEEbiography}[{\includegraphics[width=1in,height=1.25in,clip,keepaspectratio]{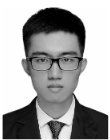}}]{Xingwu Ji}
(Graduate Student Member, IEEE) received the B.S. degree in information engineering from Xidian University, Xi'an, China, in 2019. He is currently pursuing the Ph.D. degree with Shanghai Jiao Tong University, Shanghai, China.

In 2019, he joined the Brain-Inspired Application Technology Center, Shanghai Jiao Tong University, to develop the vision brain-inspired navigation. His main areas of research interests are robotics, perception, semantic mapping.
\end{IEEEbiography}

\begin{IEEEbiography}[{\includegraphics[width=1in,height=1.25in,clip,keepaspectratio]{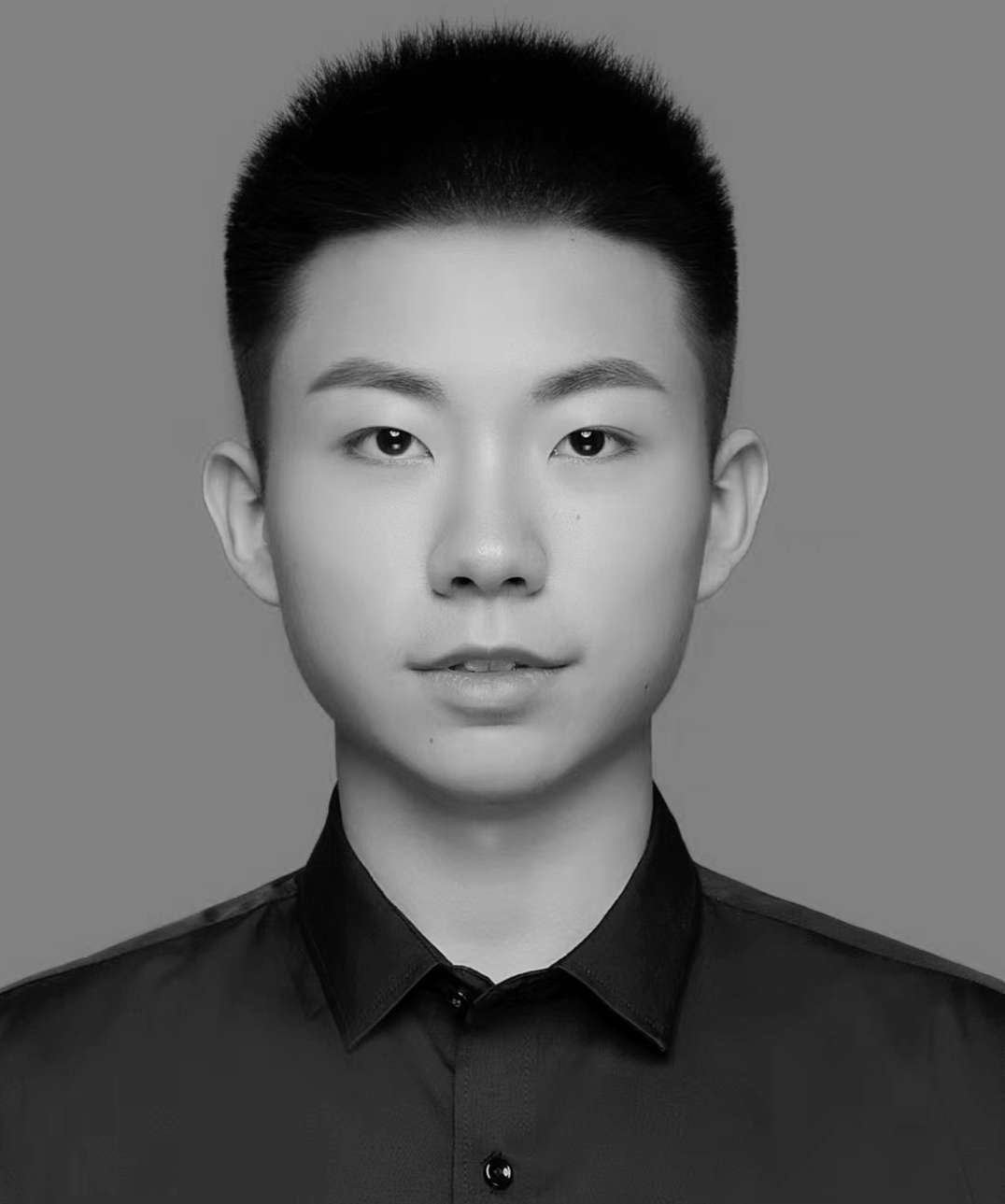}}]{Haochen Niu} 
(Graduate Student Member, IEEE) received the B.S. degree from the School of Electronics and Information Engineering, Harbin Institute of Technology, Harbin, China, in 2021. He is currently pursuing the Ph.D. degree with the School of Electronic Information and Electrical Engineering, Shanghai Jiao Tong University, Shanghai, China.

His research interests include robotics, visual SLAM and computer vision.
\end{IEEEbiography}

\begin{IEEEbiography}[{\includegraphics[width=1in,height=1.25in,clip,keepaspectratio]{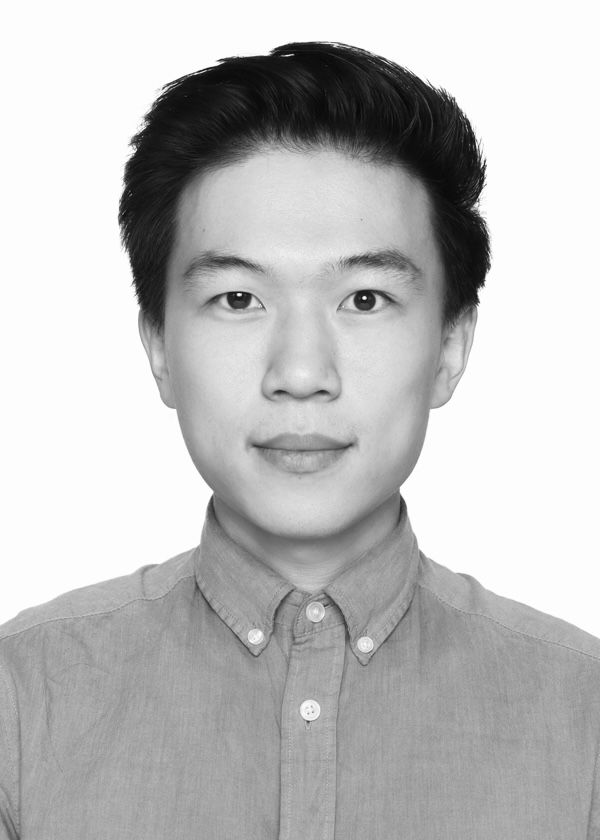}}]{Dexin Duan} received the B.E. degree from Fuzhou University, Fuzhou, China, in 2019, and the M.S. degree from the First Institute of Oceanography, Qingdao, China in 2022. He is currentlly pursuing the Ph.D. degree with the School of Electronic Information and Electrical Engineering, Shanghai Jiao Tong University, Shanghai, China.
His research interests include spiking neural networks, brain-inspired intelligence and neuromorphic computing.
\end{IEEEbiography}


\begin{IEEEbiography}[{\includegraphics[width=1in,height=1.25in,clip,keepaspectratio]{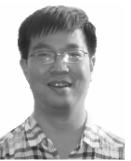}}]{Rendong Ying}
(Member, IEEE) received the B.S. degree from East China Norm University, Shanghai, China, in 1994, and the master’s and Ph.D. degrees from Shanghai Jiao Tong University, Shanghai, in 2001 and 2007, respectively, all in electronic engineering.

He is currently a professor with the Department of Electronic Engineering, Shanghai Jiao Tong University. His research area includes digital signal processing, SoC architecture, and machine thinking.
\end{IEEEbiography}

\begin{IEEEbiography}[{\includegraphics[width=1in,height=1.25in,clip,keepaspectratio]{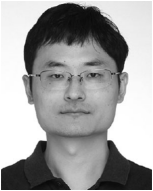}}]{Fei Wen}
(Senior Member, IEEE) received the B.S. degree from the University of Electronic Science and Technology of China (UESTC) in 2006, and the Ph.D. degree in information and communications engineering from UESTC in 2013. 

Now he is a research Associate Professor in the School of Electronic Information and Electrical Engineering at Shanghai Jiao Tong University. His main research interests are image processing, machine learning and robotics navigation. He serves as an Associate Editor of the International Journal of Robotics Research.
\end{IEEEbiography}

\begin{IEEEbiography}[{\includegraphics[width=1in,height=1.25in,clip,keepaspectratio]{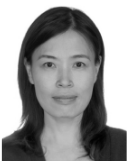}}]{Peilin Liu}
(Senior Member, IEEE) received the PhD degree from the University of Tokyo majoring in Electronic Engineering in 1998 and worked there as a research fellow in 1999. From 1999 to 2003, she worked as a senior researcher with the Central Research Institute of Fujitsu, Tokyo. 

Her research interests include low power computing architecture, application-oriented SoC design and verification, and 3D vision. She is currently a professor with Shanghai Jiao Tong University.
\end{IEEEbiography}

\vfill \eject
\end{nocolor}
\end{document}